\documentclass[letterpaper, 10 pt, conference]{ieeeconf}  

\usepackage{graphicx}
\usepackage{amsmath}
\let\labelindent\relax
\usepackage{enumitem}
\usepackage{amssymb}
\usepackage{algorithm}
\usepackage[noend]{algpseudocode}
\floatstyle{ruled}
\restylefloat{algorithm}
\usepackage{subcaption}
\usepackage[dvipsnames,table]{xcolor}
\usepackage{booktabs} 
\usepackage{multirow} 
\usepackage{array}     
\usepackage{xurl}
\usepackage{hyperref}
\usepackage{cite}
\hypersetup{
    breaklinks=true
}
\usepackage{xcolor}
\usepackage{soul}

\definecolor{profblue}{HTML}{4E79A7}
\definecolor{profgold}{HTML}{DAA520}

\newif\ifshowhighlights
\showhighlightsfalse     

\definecolor{revision}{RGB}{200,0,0}

\newenvironment{highlight}
  {\ifshowhighlights\color{revision}\fi}
  {}

\IEEEoverridecommandlockouts                              

\title{\LARGE \bf
You've Got a Golden Ticket: \\ Improving Generative Robot Policies With A Single Noise Vector
}

\author{Omkar Patil$^{1,2,*}$, Ondrej Biza$^{1}$, Thomas Weng$^{1}$, Karl Schmeckpeper$^{1}$, Wil Thomason$^{1}$, Xiaohan Zhang$^{1}$,\\
Kausik Sivakumar$^{1}$, Robin Walters$^{1,3}$, Nakul Gopalan$^{2}$, Sebastian Castro$^{1}$, Stephen Hart$^{1}$, Eric Rosen$^{1}$
\thanks{\makebox[\linewidth][l]{\parbox[t]{0.48\linewidth}{$^{1}$Robotics and AI Institute (RAI)}\hfill\parbox[t]{0.5\linewidth}{$^{2}$Arizona State University (ASU)}}}%
\thanks{\makebox[\linewidth][l]{\parbox[t]{0.48\linewidth}{$^{3}$Northeastern University}\hfill\parbox[t]{0.5\linewidth}{$^{*}$Work done while interning at RAI.}}}%
}

\begin{document}

\maketitle
\thispagestyle{empty}
\pagestyle{empty}

\begin{abstract}
Generative robot policies trained on demonstrations using behavior cloning often learn actions that are sub-optimal or misaligned with respect to the downstream task. Policy improvement approaches aim to bridge this gap and improve the cumulative reward with minimal interventions on the pre-trained policy. However, an impediment to their practical deployment is the requirement of cumbersome hyperparameter tuning specific to model or task families. In this work, we present an episodic, derivative-free latent policy improvement approach that works with little to no changes across model scales from MLPs to VLAs. We demonstrate that the performance of a pretrained, frozen diffusion or flow matching policy can be improved with respect to a downstream reward by swapping the sampling of initial noise from the prior distribution (typically isotropic Gaussian) with a well-chosen, constant initial noise input---a \emph{golden ticket}. We show the prevalence of golden tickets by improving the policy performance of $46$ out of $51$ tasks across manipulation benchmarks, with absolute improvements in success rate by up to $79\%$ for simulated tasks, and $28\%$ within $60$ search episodes for real-world tasks. Further, we find that the versatility of our approach opens up new avenues such as simultaneous policy improvement for multiple downstream rewards. Project webpage: \url{https://lottery-tickets.rai-inst.com/}.
\end{abstract}

\section{INTRODUCTION}

\begin{figure}
    \centering
    \includegraphics[width=1\linewidth]{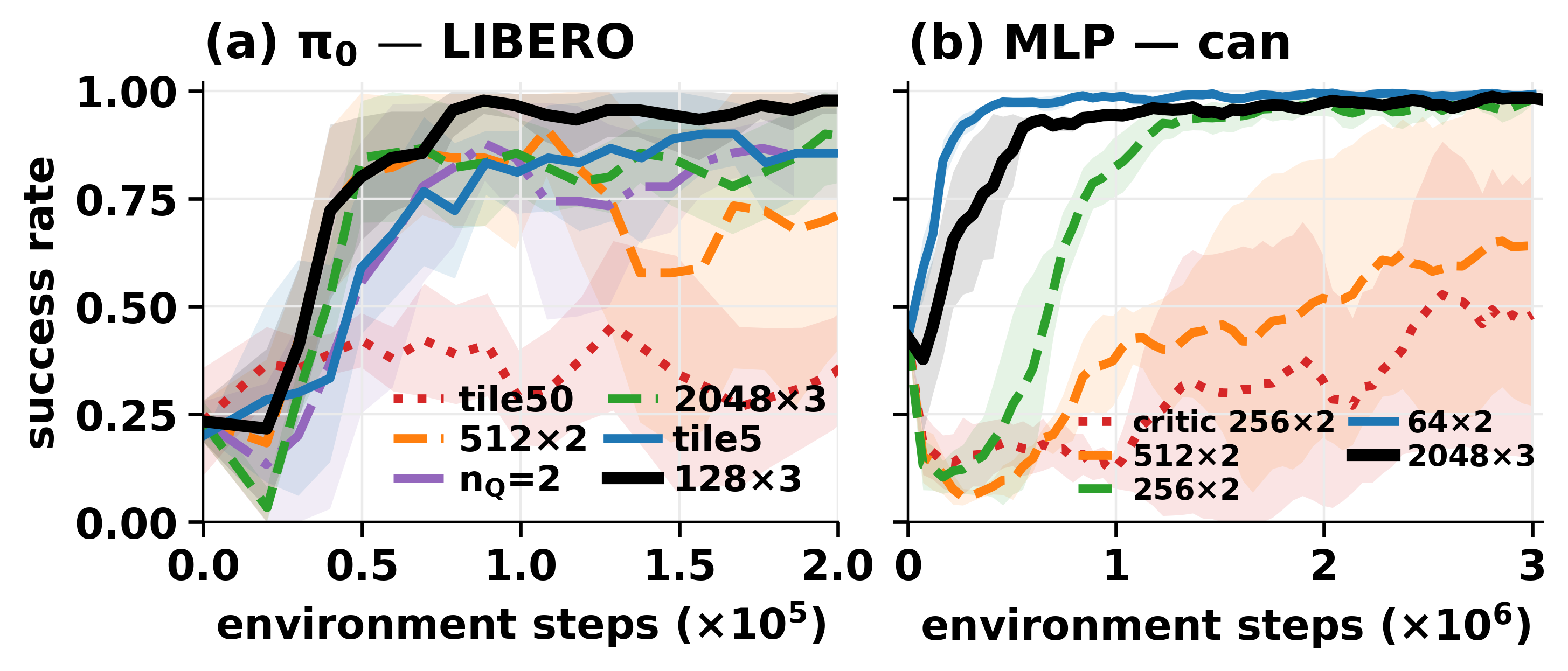}
    \caption{Model design challenges for derivative-based methods: Two plots showing only DSRL curves with different hyperparameters. Actor-critic MLP used for \texttt{Can} task $128\!\times\!3$ is not optimal on \texttt{LIBERO\#57}. (a) $\pi_0$'s larger action space needs $10$ critics $n_Q$ and noise space tiling. (b) Actor-critic MLP $64\!\times\!2$ beats the reported result (black) but $512\!\times\!2$ fails. \textit{Can policy improvement be made simpler and robust?}}
    \label{fig:dsrl-hp}
    \vspace{-0.75cm}
\end{figure}

Conditional diffusion~\cite{ho2020denoisingdiffusionprobabilisticmodels} and flow matching models~\cite{lipman2023flowmatchinggenerativemodeling} are popular approaches for learning robot control policies, because of their ability to represent high-dimensional, multi-modal action distributions.
However, these models often learn sub-optimal actions from noisy demonstrations or are not aligned with the downstream task-reward specification. Hence, generative policy improvement approaches~\cite{ren2024diffusion,wagenmaker2025steering,wang2025inferencetimepolicysteeringhuman,gu2026haf,longhini2026behavioral} aim to improve the pretrained policy's performance with respective to the downstream reward.
However, currently improving their out-of-the-box performance introduces several computational and model design challenges.
Existing methods for improving generative policies focus largely on derivative-based approaches: finetuning the pre-trained model weights~\cite{ren2024diffusion}, training an additional network~\cite{silver2018residual, wagenmaker2025steering}, or assuming specific network architectures~\cite{wang2025inferencetimepolicysteeringhuman} or training regimes~\cite{intelligence2025pi06vlalearnsexperience}.
These prior methods are computationally expensive, require complex model design choices and hyperpameter tuning (see Fig.~\ref{fig:dsrl-hp}), or require assumptions that limit the scenarios in which they can be applied.

\begin{figure*}[htbp]
    \centering
    \includegraphics[width=0.98\linewidth]{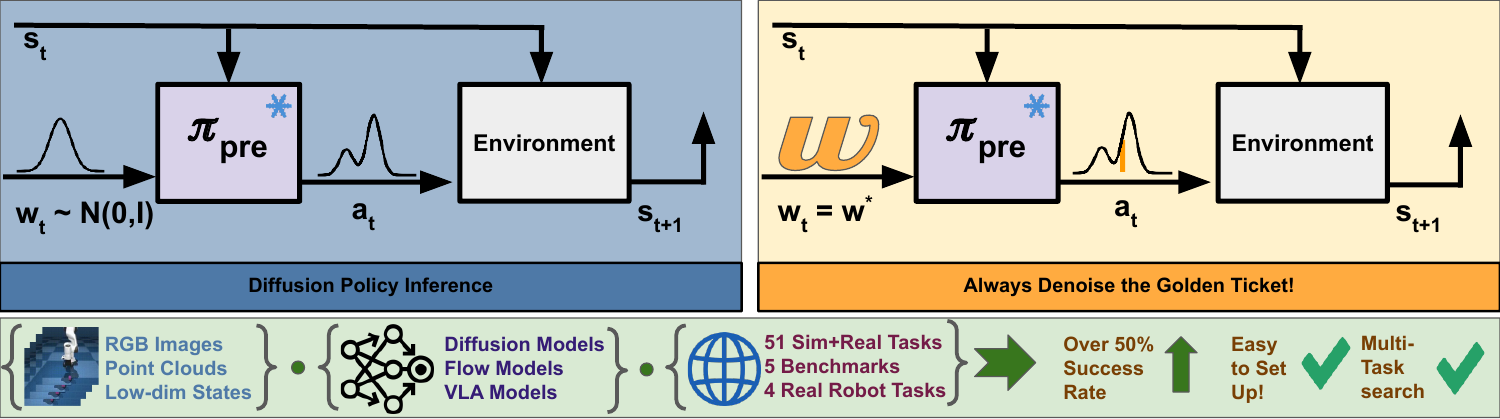}
    \caption{
        Overview of standard diffusion policy inference (left) versus our proposed inference approach of using golden tickets (right).
        Given a frozen, pretrained diffusion or flow matching policy $\pi_{\text{pre}}$, rather than sampling from a Gaussian every time an action needs to be computed, we use a constant, well-chosen initial noise vector $w$, called a golden ticket.
        We find golden tickets improve policy performance across a range of observation inputs, model architectures, and embodiments.
    }
    \vspace{-0.5cm}
    \label{fig:mainfig}
\end{figure*}

\begin{highlight}
In the fields of policy search~\cite{deisenroth2013survey} and optimization~\cite{jordana2025introduction,rubinstein1997optimization}, \emph{derivative-free} methods provide an alternative approach to derivative-based methods such as actor-critic and policy gradients. For example, techniques such as CEM~\cite{rubinstein1997optimization} and CMA-ES~\cite{hansen2001completely} have long been used for policy search in robotics~\cite{deisenroth2013survey}.
Derivative-free methods have the benefit of requiring only objective evaluations, enabling optimization over black-box, discontinuous, or otherwise non-differentiable systems~\cite{jordana2025introduction}. Moreover, derivative-free methods such as ES \cite{salimans2017evolution} and ARS~\cite{mania2018simple} have also been proposed as simpler and more robust alternatives to hyperparameter-sensitive approaches for learning control policies.
We therefore ask whether episodic and derivative-free (black-box) approaches can be used for optimizing the performance of pretrained generative robot control policies. 
While prior derivative-free policy search methods~\cite{deisenroth2013survey} directly sample the space of model parameters, we instead focus on a feature of diffusion and flow matching: selecting the initial noise vector used for the trajectory denoising process.

In this work, we demonstrate that the performance of a pretrained, frozen diffusion or flow matching policy can be improved by simply replacing the sampling of initial noise from the prior distribution with a well-chosen, constant initial noise input, which we call a \textit{golden ticket} (Fig.~\ref{fig:mainfig}).
Given a pretrained policy, a reward function, and an environment for a new downstream task, we pose finding golden tickets as a search problem, where candidate initial noise vectors, which we call
\textit{lottery tickets}, are optimized to find the ticket that maximizes the expected cumulative discounted rewards on the downstream task. 

At first glance, one might expect using a fixed noise vector for denoising to perform worse than the standard approach of sampling independent noise vectors at each step.
Our counterintuitive observation is that \textit{a single, well-chosen noise vector encodes critical information about how the policy should do the task}.
This phenomenon has been investigated in prior work on image generation~\cite{mao2024lottery, zhou2025golden, chen2024find, miao2025minimalist}, but its role in temporally conditioned generative robot policies is underexplored. 
DSRL~\cite{wagenmaker2025steering} is the closest work in this direction, proposing a derivative-based method that learns to select diffusion noise inputs via reinforcement learning.
However, while DSRL can achieve strong results, it is sensitive to hyperparameter tuning, does not scale well to large action spaces, and applies gradient-based learning to a highly non-smooth noise-selection landscape.
Although individual noise vectors are smoothly transformed through the denoising process, nearby points in the initial noise space can map to substantially different actions.
Noise selection is therefore a natural candidate for gradient-free optimization.


We propose Golden Ticket (GT) Search as a lightweight but surprisingly effective policy improvement approach that requires little-to-no hyperparameter tuning.
GT Search is intended less as a universal policy improvement algorithm than as a simple method for probing how much performance can be recovered through direct search over latent noise.
We show that searching for constant noise vectors in the policy's latent space can yield comparable performance to gradient-based methods that learn state-conditioned noise predictors, especially for higher dimensional action spaces.
GT Search keeps the pre-trained policy weights frozen, does not train an additional network, and can be applied to any diffusion or flow-matching framework without additional assumptions.
\end{highlight}

Our experimentation is designed to first empirically prove the existence and prevalence of golden tickets for a wide range of benchmarks and policies, following which we discuss the sample efficiency of searching golden tickets. On existence, we show an improvement over base policy performance in $46$ out of $51$ tasks across five simulated manipulation benchmarks, and 4 real-world tasks with a variety of pretrained policy classes. When we apply our approach on real robot hardware, we increase the relative success rate of an object picking task by $18\%$, and a piston assembly task by $28\%$, with the ticket search taking less than $150$ episodes per task. On sample-efficiency, we find that golden tickets achieve comparable sample efficiency to that of DSRL, especially with larger action spaces such as of VLAs, without requiring extensive hyperparameter tuning nor forcing the policy to tile its noise space. Finally, we also show that our method can be used for multi-reward optimization without incurring additional samples. Our contributions are:
\begin{itemize}[nosep, leftmargin=*]
    \item Golden Ticket Search, a derivative-free algorithm for finding golden tickets via  Monte-Carlo policy evaluation.
    \item Evidence of the existence, prevalence, and empirical performance of golden tickets across model architectures, observation modalities, and embodiments in 51 real and simulated tasks.
    \item Demonstration that the search for golden tickets can optimize multiple downstream reward functions (i.e., success and speed) without additional environment interaction.
    \item Open-source code for GT Search and identified golden tickets for pretrained VLA, diffusion, and flow-matching policies in three simulated manipulation benchmarks.
\end{itemize}


\section{Background}
\subsection{Markov Decision Processes.} We model robot manipulation as a Markov Decision Process (MDP).
An MDP comprises a tuple $\{S,A,T,R,p_0,\gamma\}$, where $S$ and $A$ are sets of states and actions (respectively), $T(s' \mid s, a) = \Pr(s' \mid s,a)$ is the transition dynamics, $R(s,a)$ is the reward function, $p_0 \in \triangle_{S}$ is the initial state distribution over $S$, and $\gamma$ is the discount factor.
A policy $\pi$ rolls out episodes from an initial state $s_0 \sim p_0$ by iteratively sampling an action $a_t = \pi(* \mid s_t)$ to execute at each time step $t$ to get the next state $s_{t+1} \sim T(* \mid s_t, a_t)$ and reward $r_t = R(s_t, a_t)$.
The $Q$-function $Q^{\pi}(s, a)$ for $\pi$ gives the expected discounted return of $\pi$ rolled out from a state $s$ with initial action $a$, following $\pi$ for all subsequent actions.
That is, $Q^{\pi}(s,a) = \mathbb{E}_{P^{\pi,T}(*)}[\sum_{t>0} \gamma^t R(s_t,a_t) \mid s_0 = s, a_0 = a ]$.

\subsection{Diffusion and Flow Matching.} These generative methods iteratively refine a sample from a source distribution into a desired target data distribution. We focus on the flow matching framework with a Gaussian source distribution for illustration. In flow matching, given a dataset $D = \{s_t, a_t \sim p_{\text{data}}\}_{t=1}^{n}$, the forward process gradually corrupts the action $a \in D$ over time $\tau$\footnote{We use a superscript $(\tau)$ for the forward/reverse process time to distinguish it from the MDP time subscript $t$.} by linearly interpolating it with a sample from an isotropic Gaussian: $a^{(\tau)} = (1 - \tau)\, a^{(0)} + \tau w$, where $w \sim \mathcal{N}(0,I)$ and $a^{(\tau)}$ represents the partially noised action $a^{(0)}$ at time $\tau$. Generating new samples from the data distribution $p_{\text{data}}$ requires reversing the forward process. Specifically, a denoising model $\hat{u} = \hat{u}(s, a^{(\tau)}, \tau)$ is trained to undo the forward process, and new samples from $p_{\text{data}}$ can be generated by iteratively denoising a sample $a^{(\tau)}$ with $a^{(1)} \sim \mathcal{N}(0, I)$, treating the denoising model $\hat{u}$ as a velocity field and solving the ODE $\hat{a}^{(0)} = a^{(1)} + \int_{1}^{0} \hat{u}(s, a^{(\tau)}, \tau) \, d\tau$. While diffusion and other flow variants differ in how they define the forward and reverse processes, our method is agnostic to these variations; it focuses only on making $a^{(1)}$, the initial noise vector used for inference, a constant vector rather than repeatedly sampling it from a Gaussian distribution. We consider DDIM sampling for diffusion models due to more efficient generation \cite{song2022denoisingdiffusionimplicitmodels}, which is critical in robotics.

\begin{figure*}
    \centering
    \includegraphics[width=0.98\linewidth]{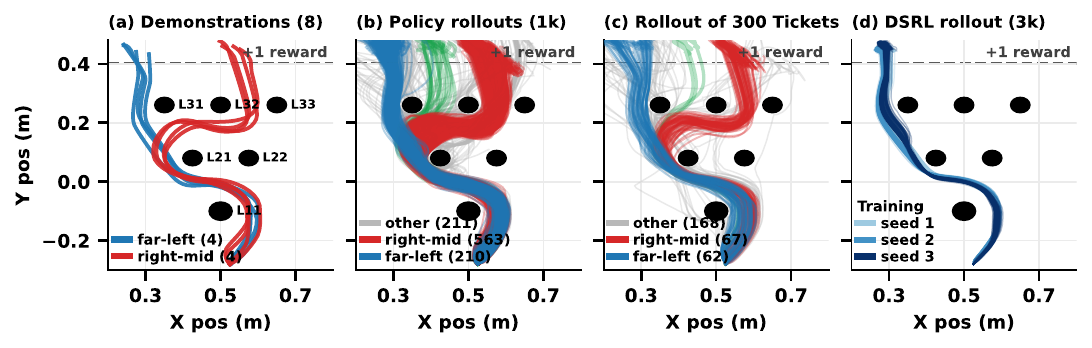}
    \caption{
        D3IL task \texttt{Avoiding-M2}: (a) Demonstrations the policy was trained on covering blue and red modes; (b) Learned policies often generate diverse behaviors (gray and green); (c) Initial noises (300 here) provide a broad mode coverage for us to choose a noise sample that optimizes the downstream reward; (d) DSRL, optimized for the blue mode. See Section~\ref{sec:viz}. \label{fig:d3il}
    }
    \vspace{-0.75cm}
\end{figure*}


\section{Related Work}

\subsection{Lottery Ticket Hypothesis for Diffusion/Flow Models.} The computer vision community has also evidenced this phenomenon, named ``lottery ticket hypothesis'' \cite{mao2024lottery}: ``randomly initialized Gaussian noise images contain special pixel blocks (winning tickets) that naturally tend to be denoised into specific content independently.''
Followup works have investigated how these ``winning tickets'' (also referred to as ``golden noises'' \cite{zhou2025golden, chen2024find, miao2025minimalist}, or ``golden tickets'' in our work) can be optimized to achieve higher text-image alignment and higher human preference when they are used instead of sampling from isotropic Gaussian noise.
However, finding golden tickets for robot control policies present additional difficulties due to the temporal nature of the problem and dimensionality of the problem space. The novelty of this work lies in empirically demonstrating the prevalence of golden tickets for robot policies and formulating their search as an episodic derivative-free policy improvement method.

\subsection{Policy Improvement via Latent Steering.} While there is a diverse range of policy improvement methods (reviewed in Appendix~\ref{app:related}), we focus on those that use latent steering, since they most closely relate to our approach.
Diffusion Steering via Reinforcement Learning (DSRL)~\cite{wagenmaker2025steering} swaps the source noise distribution with an observation-conditioned noise policy trained via reinforcement learning (RL).
However, DSRL presents two major challenges in practice (see Fig. \ref{fig:dsrl-hp}):
1) it trains actor and $2$--$10$ critic networks, which involves modeling decisions to ensure sufficient model capacity, and 2) it integrates with an RL framework to train the noise policy, which requires hyperparameter tuning~\cite{wagenmaker2025steering}.
Because golden tickets are akin to a noise policy that is observation-independent (unlike DSRL), they require no additional trainable network beyond the original policy and fewer hyperparameter adjustments. Further, DSRL fails when directly used with larger action spaces of VLAs as shown in Fig.~\ref{fig:dsrl-hp}. DSRL instead tiles the noise space, which departs from the i.i.d.\ pretraining noise and removes all temporal variation within the chunk~\cite{gu2026haf}. GT Search works directly in high dimensional noise spaces of VLAs without any such forced tiling.

\begin{highlight}
\subsection{Policy Search and Bandit-Based Evaluation}
Our search methods belong to the long line of direct policy search, which treats policy performance as a black-box objective and optimizes it from rollout returns alone, without gradients through the policy or the environment~\cite{deisenroth2013survey, salimans2017evolution, mania2018simple}.
Step-based methods perturb the action at every timestep and estimate a per-step gradient, often via the likelihood-ratio estimator~\cite{williams1992simple}. The variance of these estimates grows with both the horizon and the dimension of the action space.
Episode-based methods instead draw one perturbation per rollout in parameter space (the latent-noise space for GT) and score it by return, using finite-difference or smoothing estimators~\cite{salimans2017evolution, mania2018simple} or rank-based selection~\cite{rubinstein1997optimization, hansen2001completely}.
Evolution strategies trade sample efficiency for near-linear parallelism~\cite{salimans2017evolution}, while ARS~\cite{mania2018simple} shows that random search over simple policy classes matches gradient-based deep RL on standard continuous-control benchmarks.
Our work continues this line of inquiry, but searches the latent input to a frozen generative policy rather than the parameters of an actor. GT Search also benefits from pure-exploration
bandit~\cite{bubeck2011pure} methods. We show that sequential halving~\cite{karnin2013almost} improves the sample-efficiency of CEM, eliminating poor tickets cheaply and
concentrating evaluation on contenders.
\end{highlight}


\section{Methodology}

\begin{highlight}
\subsection{Effect of Constant Initial Noise on Policy Rollouts}
\label{sec:viz}
Randomness in generative policies can emanate from the initial noise or the sampling process used: DDPM \cite{ho2020denoisingdiffusionprobabilisticmodels} for stochastic sampling, and DDIM \cite{song2022denoisingdiffusionimplicitmodels} for deterministic sampling. DDIM has gradually evolved as the standard choice due to more efficient sampling \cite{wagenmaker2025steering}, where the initial noise remains the sole source of stochasticity in the policy. To understand the effect of constant initial noise on temporal policy rollouts and how it can serve as an instrument to optimize the downstream reward, we use the \texttt{Avoiding-M2} task (Fig. \ref{fig:d3il}) from D3IL\cite{jia2024towards}.
This task requires the robot to reach the finish line without colliding with three levels of obstacles. As visualized in Fig. \ref{fig:d3il}(b), the learned policy can often exhibit unwanted or misaligned behaviors (\textit{gray or green}).
Fig. \ref{fig:d3il}(c) shows the rollouts of 300 constant initial noise vectors, or lottery tickets. Initial noise vectors provide a broad mode coverage for us to choose a ticket that optimizes the downstream reward. While several tickets avoid \texttt{L11}, only a few go on to also avoid \texttt{L32} to be counted as \textit{red}, and even fewer form the \textit{green} mode, which lacks support in the demonstration data itself.
Hence, GTs depend on a combination of factors such as the support induced by the learned policy and the downstream reward. Interestingly, depending on the underlying search method used, once noise vectors and the corresponding rollouts are collected, they can be re-scored with different objectives (such as the green mode) to identify tickets optimizing a different or a combination of reward functions, without incurring additional environment samples. Finally, Fig. \ref{fig:d3il}(d) shows rollouts from a DSRL-augmented policy having converged to the blue mode. Later optimizing the policy for the red mode will require a new DSRL runs incurring additional environment samples, unlike random search for golden tickets.
\end{highlight}

\subsection{The Lottery Ticket Hypothesis for Robot Control.}
Our main contribution is framing the search for a golden ticket, a fixed initial noise vector, as a gradient-free approach for improving policy performance on a downstream task.
We assume we are given a pretrained, frozen diffusion or flow matching robot control policy $\pi$,
and treat this policy as a black box without access to any intermediate steps of the denoising process.
Given an MDP representing a downstream task from which we can sample episode transitions, our goal is to adapt the policy $\pi$'s behavior to maximize rewards while keeping the parameters frozen.
We propose: \textit{For a pretrained diffusion or flow matching policy and a downstream task, there exists a fixed initial noise vector $w^*$ that, used in place of sampling from the Gaussian prior, increases the policy's expected return on held-out task evaluations.}\footnote{Empirically, this holds for 46 of the 51 tasks in our benchmarks for a fixed number of trials.}

We call such a noise vector a \textit{golden ticket} and test the hypothesis empirically in Section~\ref{sec:existence}.
Searching for a golden ticket on a robot policy raises several challenges: (1) the policy acts in an environment where current actions affect future states, (2) collecting samples from that environment is costly, and (3) action and noise space of the policy may be high-dimensional. In this work, we develop a black-box search-based method using Monte-Carlo policy evaluation to address these challenges.

\subsection{Golden Tickets Search} 
We propose to use black-box \textit{search} for finding golden tickets $w^*$ with Monte-Carlo estimates of episodic-reward obtained from policy rollouts as the objective.
Formally, let $\pi_w(s)$ denote a policy whose denoising process is always initialized with noise $w \in \mathbb{R}^{H \times d_a}$, where $H$ is the horizon and $d_a$ is the per-step action dimension; the base policy $\pi(s)$ samples a fresh $w \sim \mathcal{N}(0, I)$ at every action step. We seek $w^*$ that maximizes the cumulative discounted reward obtained by $\pi_{w}$, averaged over $M$ rollouts from fixed initial states $\{s_0^{(m)}\}_{m=1}^{M}$: that is, $w^* = \operatorname*{argmax}_{w} \frac{1}{M} \sum_{m=1}^{M} \sum_{t > 0} \gamma^t R(s_t^{(m)}, \pi_w(s_t^{(m)}))$.

\begin{algorithm}
    \caption{CEM for Optimizing Initial Noise}
    \label{alg:cem_initial_noise}
    \begin{algorithmic}[1]
        \State \textbf{Input:} policy $\pi$, environments $\mathcal{M}$, population $P$, iterations $T$, elite fraction $\rho$, smoothing $\alpha$, std floor $\sigma_{\min}$, extra std $\sigma_{\text{ex}}$
        \State $\mu \gets 0$,\ $\sigma \gets 1$,\ $K \gets \max(1, \lfloor \rho P \rfloor)$,\ $\bar{R}_{\text{best}} \gets -\infty$
        \For{$t = 1$ to $T$}
            \State $\epsilon_i \gets \mu + \sigma \odot z_i$,\ $z_i \sim \mathcal{N}(0, I)$,\ $i = 1, \dots, P$
            \State $\bar{R}_i \gets \tfrac{1}{|\mathcal{M}|} \sum_{m \in \mathcal{M}} R_m(\pi, \epsilon_i)$ \Comment{roll out $\pi$ with fixed noise $\epsilon_i$}
            \State $\epsilon^* \gets \epsilon_{\arg\max_i \bar{R}_i}$ if $\max_i \bar{R}_i > \bar{R}_{\text{best}}$;\ update $\bar{R}_{\text{best}}$
            \State $\mathcal{E} \gets$ top-$K$ indices by $\bar{R}_i$
            \State $\mu \gets \alpha\, \mathrm{mean}(\epsilon_{\mathcal{E}}) + (1 - \alpha)\, \mu$
            \State $\sigma \gets \max\!\bigl(\alpha\, \mathrm{std}(\epsilon_{\mathcal{E}}) + (1 - \alpha)\, \sigma,\ \sigma_{\min}\bigr)$
            \State $\sigma \gets \sqrt{\sigma^2 + \sigma_{\text{ex}}^2}$ \Comment{if $\sigma_{\text{ex}} > 0$}
        \EndFor
        \State \Return $\epsilon^*$
    \end{algorithmic}
\end{algorithm}

In this work, we consider random search (RS)~\cite{jordana2025introduction} and the cross-entropy method (CEM)~\cite{rubinstein1997optimization} for finding golden tickets. The simplest of them, random search, samples $N$ tickets and returns the one
with the maximum average reward. CEM~\cite{rubinstein1997optimization}, shown in
Algorithm~\ref{alg:cem_initial_noise}, iteratively refines a Gaussian sampling distribution $\mathcal{N}(\mu, \operatorname{diag}(\sigma^2))$ over noise vectors, initialized at $\mu = 0$, $\sigma = 1$. Further, with sequential halving~\cite{karnin2013almost}, the environment budget $|\mathcal{M}|$ is partitioned into $R$ disjoint rounds of $M_0 = |\mathcal{M}| / R$ environments each. Unlike uniform allocation, a candidate surviving every round accumulates
$R \cdot M_0 = |\mathcal{M}|$ environment evaluations, while a candidate eliminated at round $r$ contributes only $r \cdot M_0$. While both RS and CEM only rank noise vectors, zeroth-order search (ZOS)~\cite{jordana2025introduction} instead estimates a per-coordinate gradient from scalar returns, whose variance scales as $O(Hd_a)$~\cite{duchi2015optimal}; we observe ZOS failing on some seeds.

\textbf{Evaluating Golden Tickets.} To determine if the returned ticket is ``golden'' (i.e., that it has higher average performance than sampling from the Gaussian prior), we assume there is a held-out set of initial states or environments which we use to evaluate the performance of both the best ticket and the base policy.
For a fixed compute budget, there exists a trade-off between number of tickets $N$ and number of search environments $M$ on which to evaluate those tickets. More tickets result in better coverage and potentially better performance, but risk not generalizing to held-out environments.


\section{Experiments}
Our experimental evaluation is designed to assess the lottery ticket hypothesis for robot control policies and provide empirical evidence of its effectiveness and practical utility.

\subsection{Experimental Benchmarks}
\begin{figure}[htbp]
    \centering
    \begin{subfigure}[b]{0.1\textwidth}
        \centering
        \includegraphics[width=\linewidth]{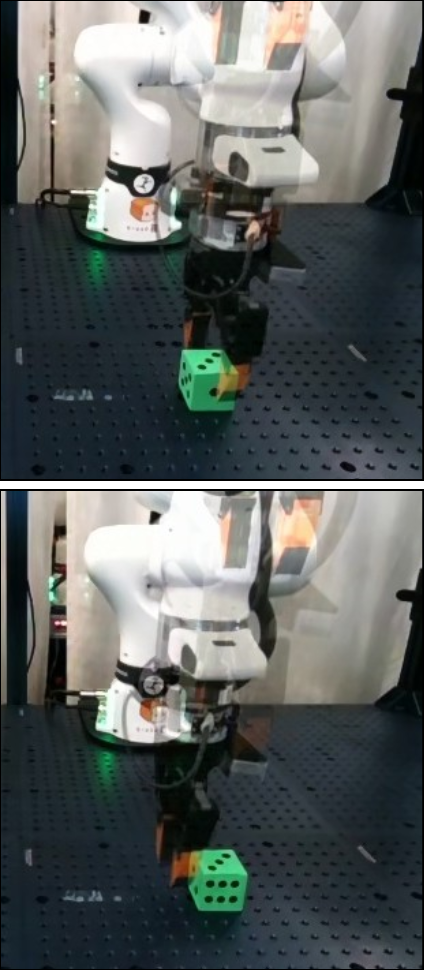}
        \caption{Pick cube}
        \label{fig:pick_cube}
    \end{subfigure}
    \begin{subfigure}[b]{0.1\textwidth}
        \centering
        \includegraphics[width=\linewidth]{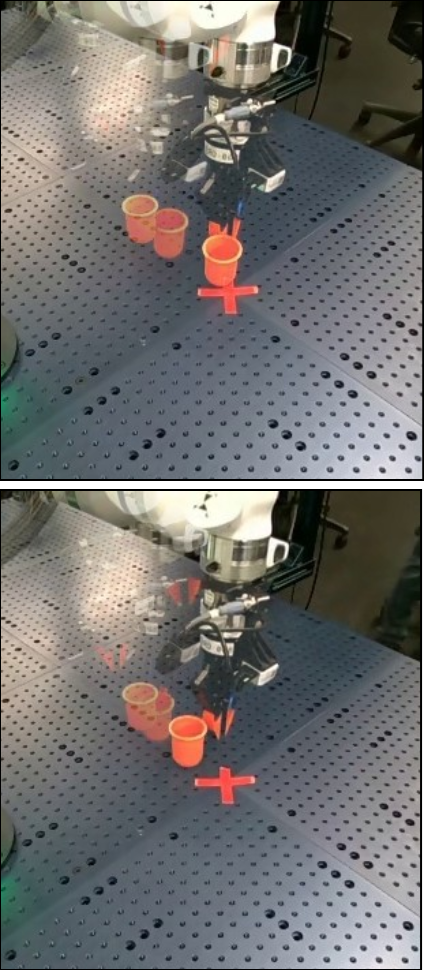}
        \caption{Push cup}
        \label{fig:push_cup}
    \end{subfigure}
    \begin{subfigure}[b]{0.1\textwidth}
        \centering
        \includegraphics[width=\linewidth]{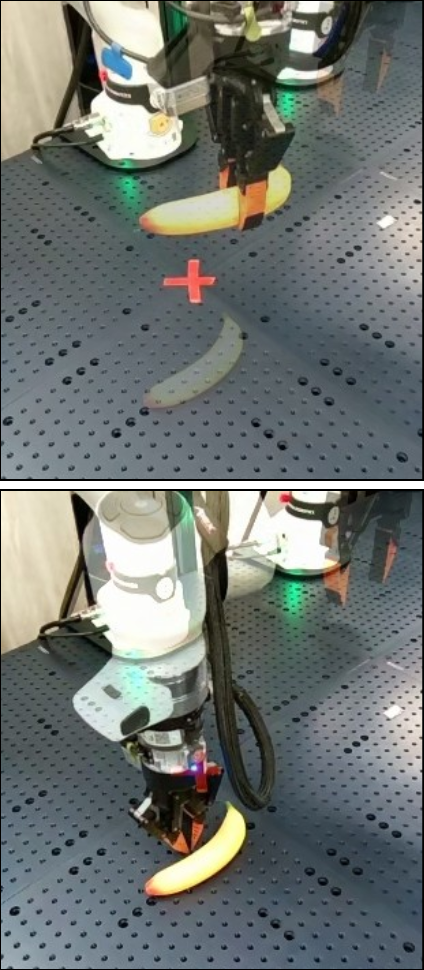}
        \caption{Banana}
        \label{fig:pick_banana}
    \end{subfigure}
    \begin{subfigure}[b]{0.1\textwidth}
        \centering
        \includegraphics[width=\linewidth]{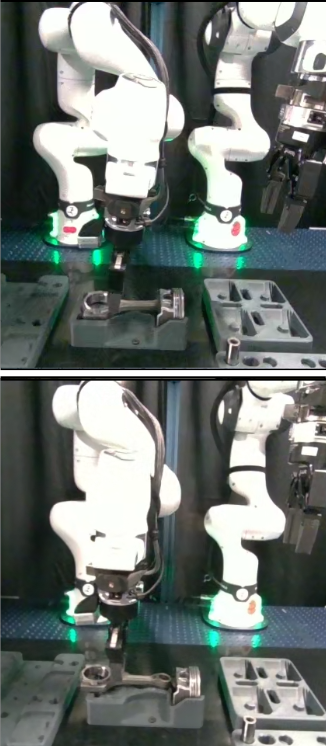}
        \caption{Piston}
        \label{fig:piston_assembly}
    \end{subfigure}
    \caption{Rollouts from diffusion policies sampling with Gaussian noise that we use in our hardware experiments. (top) An example successful rollout; (bottom) A failed rollout. }
    \label{fig:four_hardware_tasks}
    \vspace{-0.5cm}
\end{figure}

We investigate a wide range of model architectures, observation inputs, and environments. On real Franka hardware (Fig. \ref{fig:four_hardware_tasks}), we evaluate two RGB diffusion policies (piston assembly and cube picking) and two point-cloud diffusion policies (banana picking and cup pushing); all hardware experiments optimize a binary end-of-episode success signal. We also report results on the GR00T N1.5~\cite{bjorck2025gr00t} public checkpoint for 7 tasks of the \textbf{SimplerEnv} (WidowX)~\cite{li24simpler} environment, designed to reflect real world performance. Additionally, we consider 4 simulation benchmarks:
\begin{itemize}[leftmargin=1.4em,itemsep=0pt,topsep=2pt,parsep=0pt]
    \item \textbf{franka\_sim}~\cite{luo2024serl}: cube picking, flow-matching policy on low-dim state.
    \item \textbf{LIBERO}~\cite{liu2023libero}: 30 tasks (Spatial/Goal/Object) with the public SmolVLA~\cite{shukor2025smolvla} checkpoint and 3 tasks from LIBERO-90 using the public $\pi_0$ checkpoint.
    \item \textbf{robomimic}~\cite{mandlekar2021matters}: 4 tasks 
    with pretrained diffusion policies from DPPO~\cite{ren2024diffusion} on low-dim state.
    \item \textbf{DexMimicGen}~\cite{jiang2025dexmimicgen}: 5 bimanual tasks with RGB diffusion policies we train via BC.
\end{itemize}
In total, the suite covers 51 tasks and spans flow matching, BC-trained diffusion and VLA policies, with state, RGB, point-cloud, and language inputs across both simulation and real hardware. Additional benchmark details are given in Appendix~\ref{app:benchmarks}.

\subsection{Experimental Results}
\label{experiments}

\subsubsection{\textbf{Do golden tickets for robot control policies exist, and how prevalent are they? How much do they improve performance compared to sampling Gaussian noise?}}

\begin{figure*}[t]
    \centering
    \includegraphics[width=\textwidth]{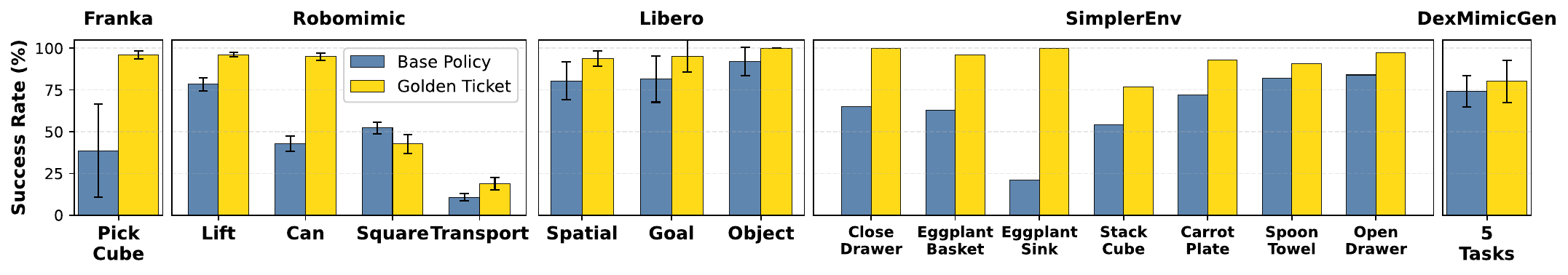}
    \caption{
        Existence and performance of golden tickets (\textbf{\textcolor{profgold}{gold}}, right) relative to the base policy (\textbf{\textcolor{profblue}{blue}}, left) on simulated benchmarks. Simply optimizing the initial noise vector can lead to strong performance gains over the base policy across benchmarks and policies. We report mean and standard deviation of success rates (details in Section \ref{experiments}). Our open-source repository will contain code and golden tickets for the first three benchmarks: franka\_sim, robomimic, and LIBERO.}
    \vspace{-0.5cm}
    \label{fig:main_results}
\end{figure*}

\begin{figure}[htbp]
    \centering
    \includegraphics[width=\linewidth]{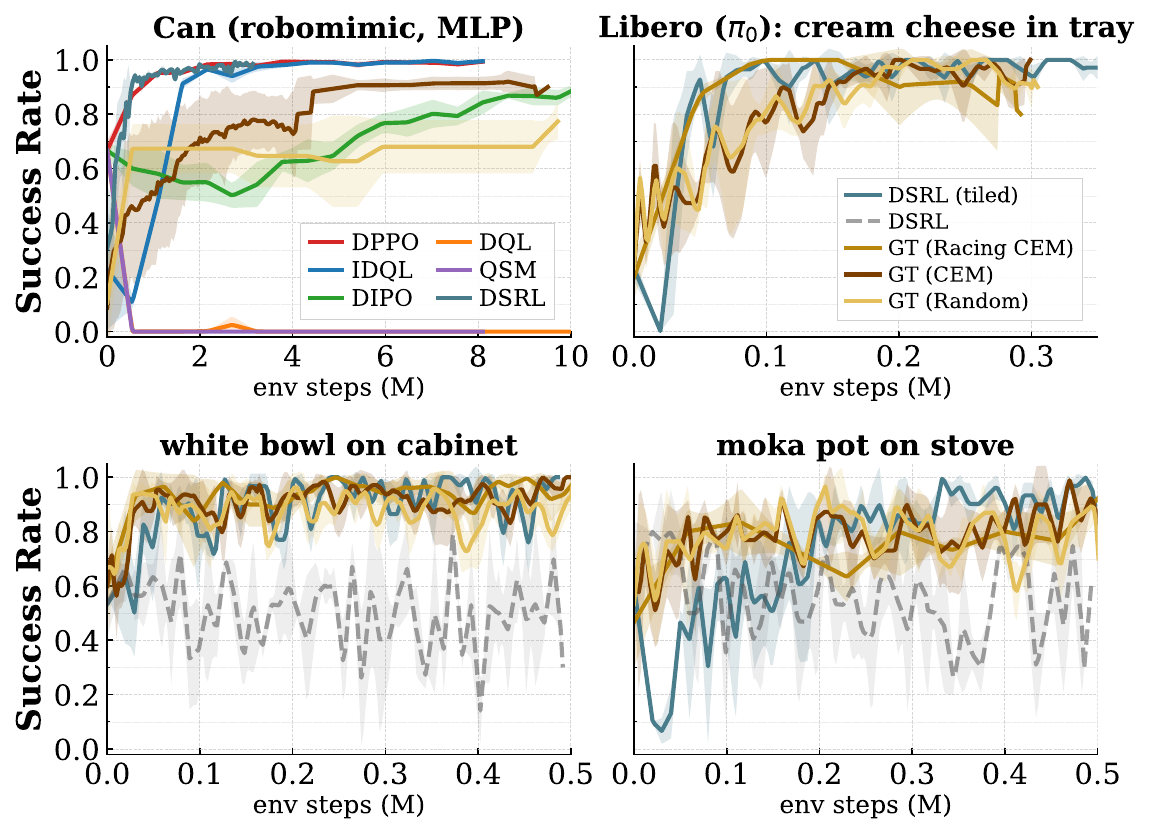}
    \caption{
    Sample efficiency comparison on \texttt{Can} with an MLP base policy (a) and three LIBERO-90 tasks with $\pi_0$ (b--d). Simply searching for constant noise vectors (GT) becomes comparable in sample efficiency to actor-critic methods at the high-dimensional noise spaces of VLA models, where DSRL on the full $1600$D noise (dashed) fails to improve on the base policy.
    }
    \vspace{-0.5cm}
    \label{fig:can-libero}
\end{figure}

\label{sec:existence}
We evaluate the existence and prevalence of golden tickets across the $51$ task--policy combinations in our suite. We run a large-scale search for golden tickets, with search budgets ranging from less than $250$ tickets with $5$ search environments each in SimplerEnv to $5000$ tickets per task with $100$ search environments each in robomimic. We find that
\textit{golden tickets outperform Gaussian noise in $46$ out of $51$ tasks, and at least match it in $49$} (Figure~\ref{fig:main_results}; per-task results in Appendix~\ref{app:existence}). Under the null hypothesis that a searched noise vector outperforms Gaussian sampling no more often than chance ($H_0\!: p \le 0.5$), the observed proportion is highly significant: one-sided binomial p-value $\approx 1.2 \times 10^{-9}$, with Wilson 95\% CI for $p$ of $[0.79, 0.96]$.
In franka\_sim cube picking, the base policy averages $38.5\%$ success and the best golden tickets average $96\%$.
In robomimic, golden tickets are found in $3$ of $4$ tasks (not \texttt{Square}), with the most extreme gain on \texttt{Can}, from $42.8\%$ to $95.0\%$.
In LIBERO, evaluating per-task golden tickets, we obtain +$13.4\%$, +$12.8\%$, and +$8\%$ improvements over the base policy on Spatial, Goal, and Object, respectively.
In DexMimicGen, golden tickets are found in $4$ of $5$ tasks; the biggest improvement is on \texttt{Box Cleanup} ($87.6\%\!\!\to\!\!97.8\%$), while \texttt{Threading} shows base $62\%$ vs.\ best ticket $60\%$.
In SimplerEnv (WidowX), the average improvement is +$30.4\%$\, with the largest gain on \texttt{put\_eggplant\_in\_sink} ($21\%\!\!\to\!\!100\%$).
On real Franka hardware, the average improvement is +$31\%$.
Beyond average improvement, we examine when golden tickets reach a high \emph{absolute} success rate: among the $34$ task--policy pairs with base success rate $\geq 71\%$, golden tickets achieve $\geq 90\%$ success in $31$ of them; among the $15$ pairs with base $\geq 87.6\%$, they achieve $\geq 97\%$ in $14$ of them. This suggests that golden tickets present a meaningful path to improving already high-performing behavior cloning policies. The significance test and the reach-threshold computation are detailed in Appendices~\ref{app:significance} and~\ref{app:reach}.

\begin{figure}[htbp]
    \centering
    \includegraphics[width=\linewidth]{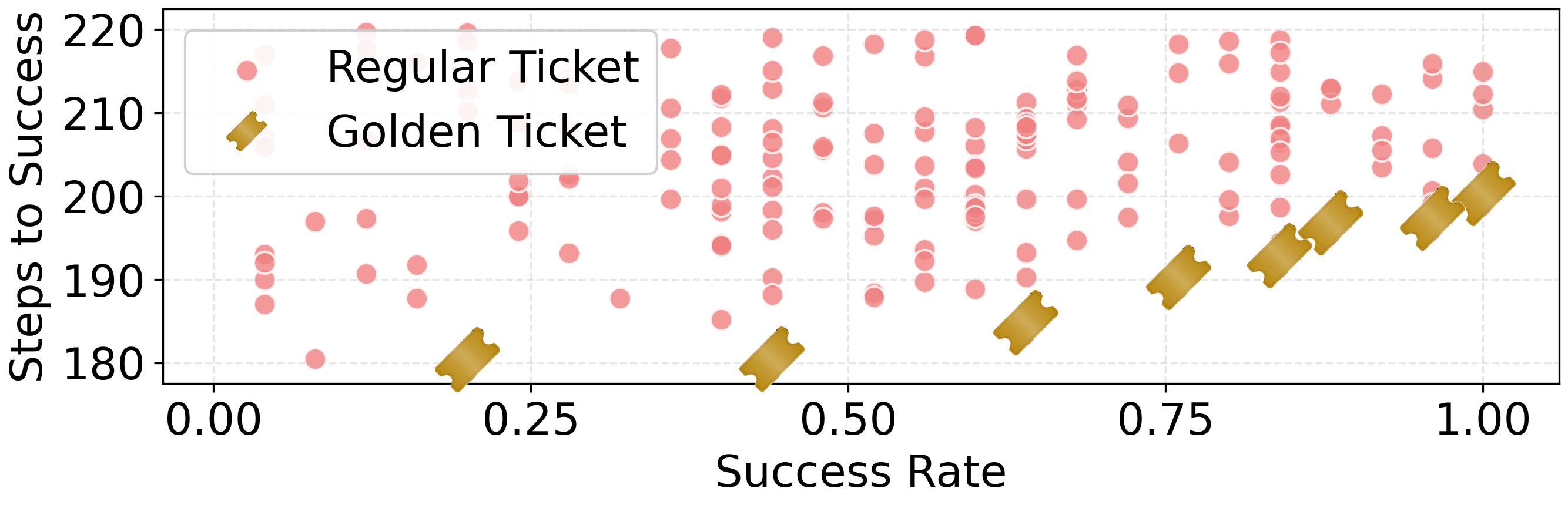}
    \caption{
    Various tickets (pink) for the franka\_sim \texttt{pick} policy, evaluated according to success rate and speed (mean length of successful episodes).
    Right is higher success rate, down is faster time to success.
    Because lottery tickets exhibit extreme differences in policy performance, a Pareto frontier is defined by tickets that are further right/down than others (represented with golden ticket icons).
    }
    \vspace{-0.5cm}
    \label{fig:pareto}
\end{figure}

\begin{highlight}
\subsubsection{\textbf{How competitive is searching for golden tickets to state-of-the-art latent steering and reinforcement learning methods?}}
\label{sec:comp-dsrl}
We compare random search and CEM (Algorithm~\ref{alg:cem_initial_noise}) against six gradient-based RL baselines including DSRL, with tuned implementations for a state-based robomimic task \texttt{Can} (Fig.~\ref{fig:can-libero}a; see \cite{ren2024diffusion} for baseline details). The other five gradient-based baselines either adapt the pretrained policy behavior (DPPO\cite{ren2024diffusion}, IDQL\cite{hansen2023idql}, DQL\cite{wang2022dql}), or use diffusion but learn from scratch (DIPO\cite{yang2023dipo}, QSM\cite{psenka2023qsm}). On \texttt{Can}, merely searching for a good fixed initial noise is competitive with gradient-based adaptation methods, even for MLP diffusion policies trained in state-based tasks -- outperforming QSM, DQL, and IDQL. A further comparison against DSRL on the five DexMimicGen tasks is given in Appendix~\ref{app:dexmimicgen-results}.

With $d_a$ as the action dimension and $H$ as the prediction horizon, as the noise space grows from $d_a \times H = 7 \times 4 = 28$D in \texttt{Can} MLP to $32 \times 50 = 1600$D in LIBERO $\pi_0$, we see DSRL falter (Fig.~\ref{fig:can-libero}). DSRL updates its actor along $\nabla_w Q$ of a critic fit by temporal-difference regression over the $d_a H$-dimensional noise space. As $d_a H$ grows, the critic's data covers a smaller fraction of this space~\cite{kleinberg2008multi}, and the actor's ascent moves toward directions where $Q$ is overestimated, which the critic then bootstraps from~\cite{fujimoto2018addressing}. DSRL tiles its noise space to reduce this regression to $d_a$ dimensions; GT search needs no such restriction.

Surprisingly, we see in Figure~\ref{fig:can-libero}b,c,d that naive random sampling of constant initial noises from the $1600$D noise space leads to comparable performance as running an actor-critic in the tiled noise space. Random search for golden tickets depends on the volume fraction of tickets in the noise space of the policy that cross the expected reward threshold. This does not directly depend on the action-space dimensionality, which is why we see GT Search perform well even for larger action spaces, such as those of VLAs. For example, we find that only $2.5\%$ of the sampled 200 tickets resulted in greater than base policy performance for $28$D \texttt{Can}, while over $50\%$ of the sampled tickets of $200$ were better than the base policy for \texttt{LIBERO\#57} $\pi_0$ with a $1600$D action space. Golden tickets benefit from being an episodic policy-improvement algorithm~\cite{mania2018simple}: they do not need to ascend along the coordinates of the noise space, which is what hurts DSRL as $d_aH$ grows.
\end{highlight}

\subsubsection{\textbf{Do lottery tickets define a Pareto frontier when given multiple objectives to optimize?}}
We investigate this in franka\_sim with a block picking task under two objectives---maximizing success rate, and maximizing speed of successful episodes---which trade off against each other (the faster the robot moves, the harder it becomes to successfully pick up the block).
We run $400$ tickets in $50$ different start states each, and evaluate each ticket on both objectives.
\textit{Golden tickets are sufficiently varied to define a Pareto frontier that balances success rate and speed for an object picking task (Fig.~\ref{fig:pareto}).}
Ticket success rates vary from ${\approx}5\%$ to nearly $100\%$, and speeds of successful episodes range from ${\approx}180$ to $220$ steps; this diversity results in a small subset of golden tickets that define a Pareto frontier, outperforming all other regular tickets to the left and above them and representing a balance between the two objectives.
This suggests a simple way to alter a robot's behavior to adapt between objectives online: collect the golden tickets on the Pareto frontier, and switch between them as desired. Appendix~\ref{app:multi-task} additionally demonstrates that golden tickets can be found for multiple tasks simultaneously.

\subsubsection{\textbf{How effective is Golden Tickets Search in practice and how do they perform on hardware?}}
On hardware, Golden Ticket Search delivers substantial gains within $1$ hour of rollout time on held-out episodes for every task:
\begin{itemize}[leftmargin=1.4em,itemsep=2pt,topsep=2pt,parsep=0pt]
    \item \texttt{block pick}: $80\%\!\!\to\!\!98\%$ ($+18\%$) on $50$ held-out episodes; $6$ tickets $\!\times\!25$ episodes $\!=\!150$ search episodes (${\approx}30$ min). The same search also surfaces a ticket succeeding only $2/50$ times, letting us steer the policy to \emph{miss} with extreme reliability.
    \item \texttt{piston assembly}: $60\%\!\!\to\!\!88\%$ ($+28\%$) on $25$ eps; $22$ tickets over $61$ search eps (${\approx}$45 mins).
    \item \texttt{banana pick}: $50\%\!\!\to\!\!68\%$ ($+18\%$) on $50$ eps; $10$ tickets $\!\times\!5$ eps $\!=\!50$ search eps (${\approx}11$ min).
    \item \texttt{cup push}: $40\%\!\!\to\!\!100\%$ ($+60\%$) on $10$ eps; $10$ tickets $\!\times\!5$ eps $\!=\!50$ search eps (${\approx}10$ min).
\end{itemize}
In line with showing improvements on public checkpoints, we also search for golden tickets on the GR00T N1.5 checkpoint for the SimplerEnv (WidowX), an environment designed to correlate closely with real world performance. Across $3$ seed trials for each task individually, we are able to improve the base VLA performance by an average of $20\%$ (with $5$ of $7$ showing substantial gains) using a search budget of less than $100$ tickets on average. We use $5$ evaluations per ticket for $4$ tasks and $10$ evaluations per ticket for $3$ tasks with relatively higher performance. Average single-task gains reach $+48\%$ on \texttt{put\_eggplant\_in\_sink} ($21\%\!\!\to\!\!69.3\%$), $+34\%$ on \texttt{close\_drawer} ($65\%\!\!\to\!\!99.2\%$), and $+28\%$ on \texttt{put\_eggplant\_in\_basket} ($63\%\!\!\to\!\!91.6\%$).


\section{Analysis and Discussion}
\subsection{\textbf{How susceptible is Golden Ticket Search to hyperparameter changes?}}
\begin{figure}[htbp]
    \centering
    \includegraphics[width=\linewidth]{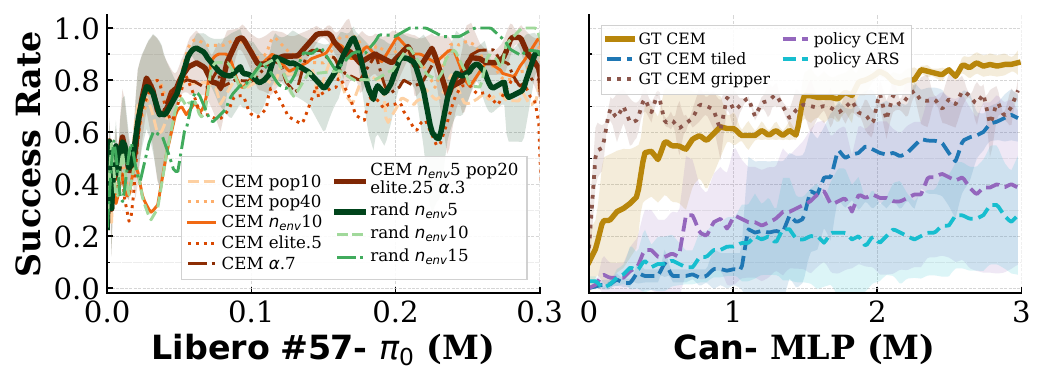}
    \caption{
    (a) Effect of hyperparameter changes on GT search for \texttt{cream cheese in tray}. (b) GT ablations: searching a tiled space leads to worse performance due to a constrained search space. Search in latent policy parameter space (akin to DSRL actor) also leads to poor performance.
    }
    \vspace{-0.25cm}
    \label{fig:can-ablation-gt-hp}
\end{figure}

As we do not learn any external network, the search for golden tickets is naturally robust to sizing the MLP and other associated hyperparameters which often result in poor performance for DSRL as shown in Fig. \ref{fig:dsrl-hp}. Random search for golden tickets is only affected by the number of environments used to form the Monte-Carlo estimate of the cumulative reward for each ticket. CEM, while more sample efficient, comes with its own set of search related hyperparameters, as does SAC for DSRL. We show the associated performance variations with hyperparameters for Golden Ticket Search in Fig. \ref{fig:can-ablation-gt-hp}(a).

\begin{highlight}
\subsection{\textbf{Wall clock time comparison for Golden Ticket Search}}
Searching for golden tickets is also far cheaper than DSRL in wall-clock time for the same env-step budget. Table~\ref{tab:dsrl-wallclock} reports the time to run $1$M environment steps on \texttt{Lift} and $2$M on \texttt{Can} (three seeds per method, one 24GB RTX3090 with a 24-core CPU): plain CEM is about $15$--$20\times$ faster than DSRL and CEM+racing about $5\times$, with random search faster still. The gap is DSRL's cost of training an actor-critic noise policy on top of the rollouts, which grows further with the $5$--$10$ critics needed to curb over-estimation~\cite{wagenmaker2025steering}; ticket search trains nothing, is independent of the observation modality, and scales purely with rollout time. Random search over tickets is also parallel, and while CEM and racing are sequential across iterations (buying sample efficiency at some wall-clock cost), they remain far faster than DSRL and gradient-free, mirroring the parallelism gains of evolution strategies~\cite{salimans2017evolution}.
\end{highlight}

\begin{table}[h]
    \centering
    \footnotesize
    \setlength{\tabcolsep}{4pt}
    \caption{Wall-clock minutes to run $1$M env steps on \texttt{Lift} and $2$M on \texttt{Can} (speed-up over DSRL in parentheses)}
    \label{tab:dsrl-wallclock}
    \begin{tabular}{l r r}
        \toprule
        \textbf{Method} & \texttt{Lift} ($1$M) & \texttt{Can} ($2$M) \\
        \midrule
        Random search & $15$ ($23\times$) & $32$ ($26\times$) \\
        Plain CEM     & $23$ ($15\times$) & $42$ ($20\times$) \\
        CEM+racing    & $67$ ($5\times$)  & $149$ ($6\times$) \\
        DSRL          & $345$              & $827$ \\
        \bottomrule
    \end{tabular}
    \vspace{-0.5cm}
\end{table}

\begin{highlight}
\subsection{\textbf{Search method ablations and bandit evaluations}}
In Fig. \ref{fig:can-ablation-gt-hp}(b), we present ablations such as searching within the subspace corresponding to the gripper dimensions, while randomly sampling the rest of noise matrix randomly. We see that the performance on searching the gripper subspace rises very quickly but plateaus below the performance of full-noise space search. Tiling the noise vector along the horizon leads to worse performance due to a smaller search space.  
Finally, we also compare against traditional policy search, but for a latent noise prediction model. We also use CEM~\cite{rubinstein1997optimization} and ARS~\cite{mania2018simple} to search over the parameters of a small policy that maps the observation to an initial noise vector, akin to the DSRL actor, which is instead learned with actor-critic. Not only does this obtain worse performance than Golden Ticket Search in the state-based setting, it is also not straightforward to adapt to VLAs, requiring a much larger network that is difficult for these methods to scale.
\end{highlight}



\begin{highlight}
\section{Limitations and Future Work}
\label{sec:futurework}
There are important limitations to our work that are addressable in future work.
\begin{itemize}[nosep, leftmargin=*]
\item Similar to DSRL, the efficacy of our method depends on the steerability of the policy~\cite{wagenmaker2025steering}. While GT Search is not directly affected by the size of the policy's action space, its performance depends on the fraction of total tickets that cross the required reward threshold. Developing a mechanistic or theoretical explanation for the performance of GT Search is an important future endeavor. 
\item Being state conditioned, a tuned implementation of DSRL, may lead to better performance. This work's contribution lies in investigating the surprising phenomenon that simply searching for a constant initial noise vector can lead to comparable performance to sophisticated gradient-based methods requiring cumbersome hyperparameter tuning.
\item Latent steering is bounded by the support of the base policy: a golden ticket can only elicit behaviors that the frozen policy is already capable of producing.
\item Finally, as with other latent policy steering approaches \cite{longhini2026behavioral}, GT Search chooses high reward modes for improving the performance, while also resulting in deterministic rollouts.   
\end{itemize}
\end{highlight}


\section{Conclusion}
Prior work on generative policy improvement has largely focused on derivative-based approaches.
We propose an episodic, derivative-free method based on searching for constant initial noises in the policy's latent space. Our approach to improving pretrained diffusion or flow-matching policies avoids 1) adjusting the original policy weights, 2) training additional neural networks, and 3) restricting the policy's noise space, as tiling does. Through experiments, we demonstrate the prevalence of golden tickets and show that simply optimizing these noise vectors can match state-of-the-art derivative-based latent policy improvement methods, especially in high-dimensional action spaces. We also release an open-source codebase with models, environments, and golden tickets for VLAs, diffusion policies, and flow-matching policies. {\footnotesize \url{https://lottery-tickets.rai-inst.com/}}
\label{sec:conclusion}
\vspace{-0.5cm}


\section*{ACKNOWLEDGMENT}
We thank Stefanie Tellex for providing valuable feedback on the manuscript.

\bibliographystyle{IEEEtran}
\bibliography{references}

@misc{ho2020denoisingdiffusionprobabilisticmodels,
  title         = {Denoising Diffusion Probabilistic Models},
  author        = {Jonathan Ho and Ajay Jain and Pieter Abbeel},
  year          = {2020},
  eprint        = {2006.11239},
  archiveprefix = {arXiv},
  primaryclass  = {cs.LG},
  url           = {https://arxiv.org/abs/2006.11239}
}

@misc{lipman2023flowmatchinggenerativemodeling,
  title         = {Flow Matching for Generative Modeling},
  author        = {Yaron Lipman and Ricky T. Q. Chen and Heli Ben-Hamu and Maximilian Nickel and Matt Le},
  year          = {2023},
  eprint        = {2210.02747},
  archiveprefix = {arXiv},
  primaryclass  = {cs.LG},
  url           = {https://arxiv.org/abs/2210.02747}
}

@article{ren2024diffusion,
  title   = {Diffusion policy policy optimization},
  author  = {Ren, Allen Z and Lidard, Justin and Ankile, Lars L and Simeonov, Anthony and Agrawal, Pulkit and Majumdar, Anirudha and Burchfiel, Benjamin and Dai, Hongkai and Simchowitz, Max},
  journal = {arXiv preprint arXiv:2409.00588},
  year    = {2024}
}

@article{wagenmaker2025steering,
  title   = {Steering Your Diffusion Policy with Latent Space Reinforcement Learning},
  author  = {Wagenmaker, Andrew and Nakamoto, Mitsuhiko and Zhang, Yunchu and Park, Seohong and Yagoub, Waleed and Nagabandi, Anusha and Gupta, Abhishek and Levine, Sergey},
  journal = {arXiv preprint arXiv:2506.15799},
  year    = {2025}
}

@article{silver2018residual,
  title   = {Residual policy learning},
  author  = {Silver, Tom and Allen, Kelsey and Tenenbaum, Josh and Kaelbling, Leslie},
  journal = {arXiv preprint arXiv:1812.06298},
  year    = {2018}
}

@article{longhini2026behavioral,
  title   = {Behavioral Mode Discovery for Fine-tuning Multimodal Generative Policies},
  author  = {Longhini, Alberta and Emukpere, David and Renders, Jean-Michel and Kim, Seungsu},
  journal = {arXiv preprint arXiv:2605.11387},
  year    = {2026}
}

@misc{wang2025inferencetimepolicysteeringhuman,
  title         = {Inference-Time Policy Steering through Human Interactions},
  author        = {Yanwei Wang and Lirui Wang and Yilun Du and Balakumar Sundaralingam and Xuning Yang and Yu-Wei Chao and Claudia Perez-D'Arpino and Dieter Fox and Julie Shah},
  year          = {2025},
  eprint        = {2411.16627},
  archiveprefix = {arXiv},
  primaryclass  = {cs.RO},
  url           = {https://arxiv.org/abs/2411.16627}
}

@misc{intelligence2025pi06vlalearnsexperience,
  title         = {$\pi^{*}_{0.6}$: a VLA That Learns From Experience},
  author        = {Physical Intelligence and Ali Amin and Raichelle Aniceto and Ashwin Balakrishna and Kevin Black and Ken Conley and Grace Connors and James Darpinian and Karan Dhabalia and Jared DiCarlo and Danny Driess and Michael Equi and Adnan Esmail and Yunhao Fang and Chelsea Finn and Catherine Glossop and Thomas Godden and Ivan Goryachev and Lachy Groom and Hunter Hancock and Karol Hausman and Gashon Hussein and Brian Ichter and Szymon Jakubczak and Rowan Jen and Tim Jones and Ben Katz and Liyiming Ke and Chandra Kuchi and Marinda Lamb and Devin LeBlanc and Sergey Levine and Adrian Li-Bell and Yao Lu and Vishnu Mano and Mohith Mothukuri and Suraj Nair and Karl Pertsch and Allen Z. Ren and Charvi Sharma and Lucy Xiaoyang Shi and Laura Smith and Jost Tobias Springenberg and Kyle Stachowicz and Will Stoeckle and Alex Swerdlow and James Tanner and Marcel Torne and Quan Vuong and Anna Walling and Haohuan Wang and Blake Williams and Sukwon Yoo and Lili Yu and Ury Zhilinsky and Zhiyuan Zhou},
  year          = {2025},
  eprint        = {2511.14759},
  archiveprefix = {arXiv},
  primaryclass  = {cs.LG},
  url           = {https://arxiv.org/abs/2511.14759}
}

@article{mania2018simple,
  title   = {Simple random search provides a competitive approach to reinforcement learning},
  author  = {Mania, Horia and Guy, Aurelia and Recht, Benjamin},
  journal = {arXiv preprint arXiv:1803.07055},
  year    = {2018}
}

@misc{song2022denoisingdiffusionimplicitmodels,
  title         = {Denoising Diffusion Implicit Models},
  author        = {Jiaming Song and Chenlin Meng and Stefano Ermon},
  year          = {2022},
  eprint        = {2010.02502},
  archiveprefix = {arXiv},
  primaryclass  = {cs.LG},
  url           = {https://arxiv.org/abs/2010.02502}
}

@inproceedings{mao2024lottery,
  title        = {The lottery ticket hypothesis in denoising: Towards semantic-driven initialization},
  author       = {Mao, Jiafeng and Wang, Xueting and Aizawa, Kiyoharu},
  booktitle    = {European Conference on Computer Vision},
  pages        = {93--109},
  year         = {2024},
  organization = {Springer}
}

@inproceedings{zhou2025golden,
  title     = {Golden noise for diffusion models: A learning framework},
  author    = {Zhou, Zikai and Shao, Shitong and Bai, Lichen and Zhang, Shufei and Xu, Zhiqiang and Han, Bo and Xie, Zeke},
  booktitle = {Proceedings of the IEEE/CVF International Conference on Computer Vision},
  pages     = {17688--17697},
  year      = {2025}
}

@inproceedings{chen2024find,
  title     = {Find: Fine-tuning initial noise distribution with policy optimization for diffusion models},
  author    = {Chen, Changgu and Yang, Libing and Yang, Xiaoyan and Chen, Lianggangxu and He, Gaoqi and Wang, Changbo and Li, Yang},
  booktitle = {Proceedings of the 32nd ACM International Conference on Multimedia},
  pages     = {6735--6744},
  year      = {2024}
}

@article{miao2025minimalist,
  title   = {A Minimalist Method for Fine-tuning Text-to-Image Diffusion Models},
  author  = {Miao, Yanting and Loh, William and Poupart, Pacal and Kothawade, Suraj},
  journal = {arXiv preprint arXiv:2506.12036},
  year    = {2025}
}

@article{duchi2015optimal,
  title     = {Optimal rates for zero-order convex optimization: The power of two function evaluations},
  author    = {Duchi, John C and Jordan, Michael I and Wainwright, Martin J and Wibisono, Andre},
  journal   = {IEEE Transactions on Information Theory},
  volume    = {61},
  number    = {5},
  pages     = {2788--2806},
  year      = {2015},
  publisher = {IEEE}
}

@inproceedings{fujimoto2018addressing,
  title     = {Addressing Function Approximation Error in Actor-Critic Methods},
  author    = {Fujimoto, Scott and van Hoof, Herke and Meger, David},
  booktitle = {Proceedings of the 35th International Conference on Machine Learning},
  pages     = {1587--1596},
  year      = {2018},
  editor    = {Dy, Jennifer and Krause, Andreas},
  volume    = {80},
  series    = {Proceedings of Machine Learning Research},
  month     = {10--15 Jul},
  publisher = {PMLR},
  url       = {https://proceedings.mlr.press/v80/fujimoto18a.html}
}

@inproceedings{kleinberg2008multi,
  title     = {Multi-armed bandits in metric spaces},
  author    = {Kleinberg, Robert and Slivkins, Aleksandrs and Upfal, Eli},
  booktitle = {Proceedings of the fortieth annual ACM symposium on Theory of computing},
  pages     = {681--690},
  year      = {2008}
}

@article{salimans2017evolution,
  title   = {Evolution strategies as a scalable alternative to reinforcement learning},
  author  = {Salimans, Tim and Ho, Jonathan and Chen, Xi and Sidor, Szymon and Sutskever, Ilya},
  journal = {arXiv preprint arXiv:1703.03864},
  year    = {2017}
}

@article{bubeck2011pure,
  title     = {Pure exploration in finitely-armed and continuous-armed bandits},
  author    = {Bubeck, S{\'e}bastien and Munos, R{\'e}mi and Stoltz, Gilles},
  journal   = {Theoretical Computer Science},
  volume    = {412},
  number    = {19},
  pages     = {1832--1852},
  year      = {2011},
  publisher = {Elsevier}
}

@inproceedings{karnin2013almost,
  title        = {Almost optimal exploration in multi-armed bandits},
  author       = {Karnin, Zohar and Koren, Tomer and Somekh, Oren},
  booktitle    = {International conference on machine learning},
  pages        = {1238--1246},
  year         = {2013},
  organization = {PMLR}
}

@article{rubinstein1997optimization,
  title     = {Optimization of computer simulation models with rare events},
  author    = {Rubinstein, Reuven Y},
  journal   = {European Journal of Operational Research},
  volume    = {99},
  number    = {1},
  pages     = {89--112},
  year      = {1997},
  publisher = {Elsevier}
}

@article{jordana2025introduction,
  title   = {An Introduction to Zero-Order Optimization Techniques for Robotics},
  author  = {Jordana, Armand and Zhang, Jianghan and Amigo, Joseph and Righetti, Ludovic},
  journal = {arXiv preprint arXiv:2506.22087},
  year    = {2025}
}

@article{bjorck2025gr00t,
  title   = {Gr00t n1: An open foundation model for generalist humanoid robots},
  author  = {Bjorck, Johan and Casta{\~n}eda, Fernando and Cherniadev, Nikita and Da, Xingye and Ding, Runyu and Fan, Linxi and Fang, Yu and Fox, Dieter and Hu, Fengyuan and Huang, Spencer and others},
  journal = {arXiv preprint arXiv:2503.14734},
  year    = {2025}
}

@article{li24simpler,
  title   = {Evaluating Real-World Robot Manipulation Policies in Simulation},
  author  = {Xuanlin Li and Kyle Hsu and Jiayuan Gu and Karl Pertsch and Oier Mees and Homer Rich Walke and Chuyuan Fu and Ishikaa Lunawat and Isabel Sieh and Sean Kirmani and Sergey Levine and Jiajun Wu and Chelsea Finn and Hao Su and Quan Vuong and Ted Xiao},
  journal = {arXiv preprint arXiv:2405.05941},
  year    = {2024}
}

@misc{luo2024serl,
  title         = {SERL: A Software Suite for Sample-Efficient Robotic Reinforcement Learning},
  author        = {Jianlan Luo and Zheyuan Hu and Charles Xu and You Liang Tan and Jacob Berg and Archit Sharma and Stefan Schaal and Chelsea Finn and Abhishek Gupta and Sergey Levine},
  year          = {2024},
  eprint        = {2401.16013},
  archiveprefix = {arXiv},
  primaryclass  = {cs.RO}
}

@article{shukor2025smolvla,
  title   = {Smolvla: A vision-language-action model for affordable and efficient robotics},
  author  = {Shukor, Mustafa and Aubakirova, Dana and Capuano, Francesco and Kooijmans, Pepijn and Palma, Steven and Zouitine, Adil and Aractingi, Michel and Pascal, Caroline and Russi, Martino and Marafioti, Andres and others},
  journal = {arXiv preprint arXiv:2506.01844},
  year    = {2025}
}

@article{mandlekar2021matters,
  title   = {What matters in learning from offline human demonstrations for robot manipulation},
  author  = {Mandlekar, Ajay and Xu, Danfei and Wong, Josiah and Nasiriany, Soroush and Wang, Chen and Kulkarni, Rohun and Fei-Fei, Li and Savarese, Silvio and Zhu, Yuke and Mart{\'\i}n-Mart{\'\i}n, Roberto},
  journal = {arXiv preprint arXiv:2108.03298},
  year    = {2021}
}

@inproceedings{jiang2025dexmimicgen,
  title        = {Dexmimicgen: Automated data generation for bimanual dexterous manipulation via imitation learning},
  author       = {Jiang, Zhenyu and Xie, Yuqi and Lin, Kevin and Xu, Zhenjia and Wan, Weikang and Mandlekar, Ajay and Fan, Linxi Jim and Zhu, Yuke},
  booktitle    = {2025 IEEE International Conference on Robotics and Automation (ICRA)},
  pages        = {16923--16930},
  year         = {2025},
  organization = {IEEE}
}

@article{liu2023libero,
  title   = {Libero: Benchmarking knowledge transfer for lifelong robot learning},
  author  = {Liu, Bo and Zhu, Yifeng and Gao, Chongkai and Feng, Yihao and Liu, Qiang and Zhu, Yuke and Stone, Peter},
  journal = {Advances in Neural Information Processing Systems},
  volume  = {36},
  pages   = {44776--44791},
  year    = {2023}
}

@article{williams1992simple,
  title     = {Simple statistical gradient-following algorithms for connectionist reinforcement learning},
  author    = {Williams, Ronald J},
  journal   = {Machine learning},
  volume    = {8},
  number    = {3},
  pages     = {229--256},
  year      = {1992},
  publisher = {Springer}
}

@article{hansen2023idql,
  title   = {Idql: Implicit q-learning as an actor-critic method with diffusion policies},
  author  = {Hansen-Estruch, Philippe and Kostrikov, Ilya and Janner, Michael and Kuba, Jakub Grudzien and Levine, Sergey},
  journal = {arXiv preprint arXiv:2304.10573},
  year    = {2023}
}

@inproceedings{jia2024towards,
  title     = {Towards diverse behaviors: A benchmark for imitation learning with human demonstrations},
  author    = {Jia, Xiaogang and Blessing, Denis and Jiang, Xinkai and Reuss, Moritz and Donat, Atalay and Lioutikov, Rudolf and Neumann, Gerhard},
  booktitle = {International Conference on Learning Representations},
  volume    = {2024},
  pages     = {41666--41686},
  year      = {2024}
}

@article{gu2026haf,
  title   = {HAF: Adapting Generalist VLAs to Humanoid Whole-Body Loco-manipulation via Hierarchical Action Flow and Spectral Latent RL},
  author  = {Gu, Langzhe and Hou, Chengkai and Li, Meng and Wang, Xinhua and Liu, Jiaming and Lv, Xinyuan and Zhang, Bowei and Bai, Shuanghao and Li, Guangrun and He, Jingyang and others},
  journal = {arXiv preprint arXiv:2608.16837},
  year    = {2026}
}

@article{psenka2023qsm,
  title   = {Learning a diffusion model policy from rewards via q-score matching},
  author  = {Psenka, Michael and Escontrela, Alejandro and Abbeel, Pieter and Ma, Yi},
  journal = {arXiv preprint arXiv:2312.11752},
  year    = {2023}
}

@article{deisenroth2013survey,
  title     = {A survey on policy search for robotics},
  author    = {Deisenroth, Marc Peter and Neumann, Gerhard and Peters, Jan},
  journal   = {Foundations and Trends{\textregistered} in Robotics},
  volume    = {2},
  number    = {1-2},
  pages     = {1--142},
  year      = {2013},
  publisher = {Emerald Publishing Limited}
}

@article{hansen2001completely,
  title     = {Completely derandomized self-adaptation in evolution strategies},
  author    = {Hansen, Nikolaus and Ostermeier, Andreas},
  journal   = {Evolutionary computation},
  volume    = {9},
  number    = {2},
  pages     = {159--195},
  year      = {2001},
  publisher = {MIT Press}
}

@article{yang2023dipo,
  title   = {Policy representation via diffusion probability model for reinforcement learning},
  author  = {Yang, Long and Huang, Zhixiong and Lei, Fenghao and Zhong, Yucun and Yang, Yiming and Fang, Cong and Wen, Shiting and Zhou, Binbin and Lin, Zhouchen},
  journal = {arXiv preprint arXiv:2305.13122},
  year    = {2023}
}

@article{wang2022dql,
  title   = {Diffusion policies as an expressive policy class for offline reinforcement learning},
  author  = {Wang, Zhendong and Hunt, Jonathan J and Zhou, Mingyuan},
  journal = {arXiv preprint arXiv:2208.06193},
  year    = {2022}
}

@article{chi2023diffusion,
  title     = {Diffusion policy: Visuomotor policy learning via action diffusion},
  author    = {Chi, Cheng and Xu, Zhenjia and Feng, Siyuan and Cousineau, Eric and Du, Yilun and Burchfiel, Benjamin and Tedrake, Russ and Song, Shuran},
  journal   = {The International Journal of Robotics Research},
  pages     = {02783649241273668},
  year      = {2023},
  publisher = {SAGE Publications Sage UK: London, England}
}

@article{du2025dynaguide,
  title   = {DynaGuide: Steering Diffusion Polices with Active Dynamic Guidance},
  author  = {Du, Maximilian and Song, Shuran},
  journal = {arXiv preprint arXiv:2506.13922},
  year    = {2025}
}

@article{hu2021lora,
  author     = {Edward J. Hu and
                Yelong Shen and
                Phillip Wallis and
                Zeyuan Allen{-}Zhu and
                Yuanzhi Li and
                Shean Wang and
                Weizhu Chen},
  title      = {LoRA: Low-Rank Adaptation of Large Language Models},
  journal    = {CoRR},
  volume     = {abs/2106.09685},
  year       = {2021},
  url        = {https://arxiv.org/abs/2106.09685},
  eprinttype = {arXiv},
  eprint     = {2106.09685},
  bibsource  = {dblp computer science bibliography, https://dblp.org}
}

@article{lu2025vla,
  title   = {Vla-rl: Towards masterful and general robotic manipulation with scalable reinforcement learning},
  author  = {Lu, Guanxing and Guo, Wenkai and Zhang, Chubin and Zhou, Yuheng and Jiang, Haonan and Gao, Zifeng and Tang, Yansong and Wang, Ziwei},
  journal = {arXiv preprint arXiv:2505.18719},
  year    = {2025}
}

@misc{luo2025empiricalstudycatastrophicforgetting,
  title         = {An Empirical Study of Catastrophic Forgetting in Large Language Models During Continual Fine-tuning},
  author        = {Yun Luo and Zhen Yang and Fandong Meng and Yafu Li and Jie Zhou and Yue Zhang},
  year          = {2025},
  eprint        = {2308.08747},
  archiveprefix = {arXiv},
  primaryclass  = {cs.CL},
  url           = {https://arxiv.org/abs/2308.08747}
}

@article{nakamoto2024steering,
  title   = {Steering your generalists: Improving robotic foundation models via value guidance},
  author  = {Nakamoto, Mitsuhiko and Mees, Oier and Kumar, Aviral and Levine, Sergey},
  journal = {arXiv preprint arXiv:2410.13816},
  year    = {2024}
}

@inproceedings{qi2017pointnet,
  title     = {Pointnet: Deep learning on point sets for 3d classification and segmentation},
  author    = {Qi, Charles R and Su, Hao and Mo, Kaichun and Guibas, Leonidas J},
  booktitle = {Proceedings of the IEEE conference on computer vision and pattern recognition},
  pages     = {652--660},
  year      = {2017}
}

@inproceedings{tan2019efficientnet,
  title        = {Efficientnet: Rethinking model scaling for convolutional neural networks},
  author       = {Tan, Mingxing and Le, Quoc},
  booktitle    = {International conference on machine learning},
  pages        = {6105--6114},
  year         = {2019},
  organization = {PMLR}
}

@article{hendrycks2016gaussian,
  title   = {Gaussian Error Linear Units (Gelus)},
  author  = {Hendrycks, D},
  journal = {arXiv preprint arXiv:1606.08415},
  year    = {2016}
}

@article{kingma2014adam,
  title   = {Adam: A method for stochastic optimization},
  author  = {Kingma, Diederik P},
  journal = {arXiv preprint arXiv:1412.6980},
  year    = {2014}
}

@inproceedings{ronneberger2015u,
  title        = {U-net: Convolutional networks for biomedical image segmentation},
  author       = {Ronneberger, Olaf and Fischer, Philipp and Brox, Thomas},
  booktitle    = {Medical image computing and computer-assisted intervention--MICCAI 2015: 18th international conference, Munich, Germany, October 5-9, 2015, proceedings, part III 18},
  pages        = {234--241},
  year         = {2015},
  organization = {Springer}
}

\appendices

\section{Related Work: Robot Policy Improvement Methods}
\label{app:related}
\label{app:robot-policy-improvement}
One class of approaches to improving pretrained policies involves directly finetuning the model parameters (or applying model adaption methods like Low-Rank Adaptation (LoRA)~\cite{hu2021lora}), using either supervised \cite{chi2023diffusion} or reinforcement learning (RL) losses \cite{lu2025vla}.
These approaches apply the same pretraining techniques to improve policies, but can be challenging to use, since adjusting policy weights can lead to catastrophic forgetting~\cite{luo2025empiricalstudycatastrophicforgetting} and is computationally hard for large models like VLAs. While some of these methods focus on RL training of auto-regressive models, more directly related to our work is DPPO~\cite{ren2024diffusion}, which backpropagates PPO gradients through the diffusion sampling process by treating an environment episode as an MDP that interleaves denoising and environment steps.

Another class of approaches learns a new model that adjusts policy output, such as residual policy learning \cite{silver2018residual} and latent steering \cite{wagenmaker2025steering}.
These methods avoid adjusting pretrained policy weights, which helps preserve the original behavior, but require carefully designing/training new neural modules with sufficient model capacity to avoid acting as bottlenecks \cite{tan2019efficientnet}.
A last class of approaches involves making additional training/inference-time assumptions (e.g., inference-time steering \cite{wang2025inferencetimepolicysteeringhuman}, classifier-free guidance \cite{intelligence2025pi06vlalearnsexperience}, world models \cite{du2025dynaguide}, value-guidance \cite{nakamoto2024steering}).
These methods avoid adjusting the pretrained policy weights or learning additional neural modules for policy improvement, but they do not apply to all pretrained models (e.g., classifier-free guidance assumes a particular training regime) or assume extra models are available (e.g., access to a world model to use dynamics for guidance, or a user to provide corrective behaviors).

\section{Additional Details on Benchmarks}
\label{app:benchmarks}
\begin{figure*}[t]
    \centering
    \includegraphics[width=\linewidth]{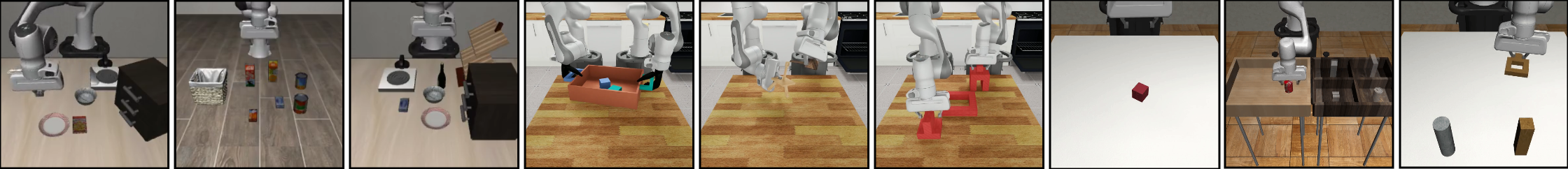}
    \caption{Images from some of our simulated benchmarks: (1-3) LIBERO Object, Spatial, and Goal task suites, (4-6) DexMimicGen \texttt{Tray Lift}, \texttt{Threading}, and \texttt{Piece Assembly} tasks, and (7-9) robomimic \texttt{Lift}, \texttt{Can} and \texttt{Square}.}
    \label{fig:tasks}
    \vspace{-0.5cm}
\end{figure*}

Table~\ref{tab:benchmarks} summarizes the benchmark suite: the policy or public checkpoint evaluated on each benchmark, its observation and action spaces, and the tasks used. ``Ours'' indicates a policy trained internally. We further describe the policies trained in our work and our real robot setup below. Figure~\ref{fig:tasks} shows sample images from the simulated benchmarks.

\begin{table*}[t]
    \centering
    \footnotesize
    \caption{Summary of experimental benchmarks. ``Ours'' indicates a policy trained internally. EE = end-effector, proprio = proprioception, PCD = point cloud.}
    \label{tab:benchmarks}
    \begin{tabular}{@{}>{\raggedright\arraybackslash}p{2.4cm}>{\raggedright\arraybackslash}p{4.5cm}>{\raggedright\arraybackslash}p{3.3cm}>{\raggedright\arraybackslash}p{6.1cm}@{}}
        \toprule
        \textbf{Benchmark} & \textbf{Policy / checkpoint} & \textbf{Obs / Action} & \textbf{Tasks} \\
        \midrule
        franka\_sim~\cite{luo2024serl} & Flow-matching MLP (open-sourced) & state / EE-pos + gripper & \texttt{pick\_cube} \\
        robomimic~\cite{mandlekar2021matters} & DPPO diffusion \newline {\scriptsize\url{github.com/irom-princeton/dppo}} & state / EE-pose & \texttt{Lift}, \texttt{Can}, \texttt{Square}, \texttt{Transport} \\
        LIBERO~\cite{liu2023libero} & SmolVLA~\cite{shukor2025smolvla} \newline {\scriptsize\url{huggingface.co/HuggingFaceVLA/smolvla_libero}} & $2\times$RGB + state + lang / $32\!\times\!50$ chunk & $30$ tasks across Object, Goal, Spatial ($10$ each) \\
        SimplerEnv (WidowX)~\cite{li24simpler} & GR00T N1.5~\cite{bjorck2025gr00t} \newline {\scriptsize\url{github.com/NVIDIA/Isaac-GR00T/tree/n1.5-release}} & RGB + lang / EE & \texttt{close\_drawer}, \texttt{open\_drawer}, \texttt{put\_eggplant\_in\_basket}, \texttt{put\_eggplant\_in\_sink}, \texttt{stack\_cube}, \texttt{carrot\_on\_plate}, \texttt{spoon\_on\_towel} \\
        DexMimicGen~\cite{jiang2025dexmimicgen} & U-Net diffusion (ours) & $3\times$RGB + proprio / bimanual EE & \texttt{Drawer Cleanup}, \texttt{Tray Lift}, \texttt{Threading}, \texttt{Box Cleanup}, \texttt{Piece Assembly} \\
        Franka HW (RGB) & U-Net diffusion (ours) & $3\times$RGB + proprio / EE & \texttt{cube\_pick}, \texttt{piston\_assembly} \\
        Franka HW (PCD) & U-Net diffusion (ours) & PCD + proprio / EE & \texttt{banana\_pick}, \texttt{cup\_push} \\
        \bottomrule
    \end{tabular}
\end{table*}

\subsubsection{Flow matching policy in franka\_sim}
We use the franka\_sim MuJoCo environment that was originally released in \cite{luo2024serl}.
The task involves a single-arm Franka robot picking a cube; the cube spawns randomly within bounds of $[-0.25, 0.25]$~m (x) and $[0.25, 0.55]$~m (y) on the ground in front of the robot (roughly a $\frac{1}{2}$~m$^2$ region), and the goal is to lift it off the ground.
We train a flow matching policy that takes as input the low-dimensional state of the cube and robot.
The policy is a $4$-layer MLP with GELU activations \cite{hendrycks2016gaussian} and $256$ hidden units per layer, and outputs an $8$-step action chunk where each step contains a 3D end-effector position and a 1D gripper state (total chunk dimension $8\times 4 = 32$).
We collect $1000$ demonstrations using a task-and-motion planning heuristic, and train $4$ model checkpoints for $100$ epochs with a batch size of $20$, using the Adam optimizer \cite{kingma2014adam} with a learning rate of $0.001$.

\subsubsection{RGB diffusion policy in DexMimicGen}
We use the DexMimicGen benchmark \cite{jiang2025dexmimicgen}, which involves five bimanual tasks with Franka arms: \texttt{Drawer Cleanup}, \texttt{Tray Lift}, \texttt{Threading}, \texttt{Box Cleanup}, and \texttt{Piece Assembly}.
For each of the $5$ tasks we train a separate diffusion policy with a U-Net backbone \cite{ronneberger2015u}, a ResNet-18 encoder for the RGB images, and an MLP for the robot proprioception data.
The policy takes in $3$ RGB images and the end-effector position, quaternion, and gripper state for both arms.

\subsubsection{Franka hardware - RGB diffusion policy}
We use Franka hardware to conduct real-world experiments: the Franka Research 3 arm is equipped with a Robotiq 2F-85 gripper, with RealSense D435 cameras for the two static external cameras and a D405 for the wrist camera.
We train RGB diffusion policies (one wrist camera, two static external cameras, and proprioception data) via behavior cloning on a block picking task(Figure \ref{fig:pick_cube}) and piston-assembly task.
Both policies uses a standard diffusion policy architecture with a U-Net backbone and ResNet-18 image encoders.
We optimize a binary success signal as the reward function: for block picking, it is whether the cube is lifted off the table, and for piston assembly, it is based on whether the piston head and rod are properly mated for subsequent pin insertion.

\subsubsection{Franka hardware - Pointcloud diffusion policy}
We use the same Franka hardware described above, but with two pointcloud policies for two separate tasks: (1) picking a banana (Figure \ref{fig:pick_banana}) and (2) pushing a cup to the center of the table (Figure \ref{fig:push_cup}).
For the pointcloud policies, we use the same U-Net backbone as in the RGB setup but replace the image encoder with a PointNet encoder \cite{qi2017pointnet} for the pointcloud.
We use $2$ calibrated extrinsic cameras to generate a single fused pointcloud, and remove all points that are at or below the surface of the table.
We again optimize a binary success signal as the reward function.

\section{Existence of Golden Tickets}
\label{app:existence}
\textbf{Golden tickets outperform Gaussian noise in $46$ out of $51$ tasks, and at least match it in $49$.} Table~\ref{tab:existence} lists, for every task--policy pair, the held-out success rate of the base policy and of the best golden ticket, together with the per-task search budget and search method. Overall, the results indicate that the lottery ticket hypothesis is often true, in the sense that golden tickets can likely be found that improve over the base policy. Table~\ref{tab:simpler_seeds} breaks the SimplerEnv results down by search seed.

\begin{table*}[t]
    \centering
    \footnotesize
    \caption{SimplerEnv (WidowX) results over three search seeds. \textbf{Gold}: best held-out success rate (\%, $300$ episodes) in each seed's CEM run; $N$: CEM candidates evaluated in each run.}
    \label{tab:simpler_seeds}
    \setlength{\tabcolsep}{10pt}
    \begin{tabular}{@{}lrrrr@{}}
        \toprule
        \textbf{Task} & \textbf{Base} & \textbf{Avg.\ Gold} & \textbf{Gold} & $N$ \\
        \midrule
        \texttt{close\_drawer} & $65$ & $99.2$ & $99.0/98.7/100.0$ & $11/16/6$ \\
        \texttt{put\_eggplant\_in\_basket} & $63$ & $91.7$ & $96.0/93.3/85.7$ & $16/32/14$ \\
        \texttt{put\_eggplant\_in\_sink} & $21$ & $69.3$ & $67.7/64.0/76.3$ & $270/27/100$ \\
        \texttt{stack\_cube} & $54$ & $71.3$ & $71.7/73.7/68.7$ & $230/69/27$ \\
        \texttt{carrot\_on\_plate} & $72$ & $74.3$ & $78.7/51.3/93.0$ & $150/106/81$ \\
        \texttt{spoon\_on\_towel} & $82$ & $79.9$ & $77.3/72.3/90.0$ & $230/53/66$ \\
        \texttt{open\_drawer} & $84$ & $95.7$ & $96.0/97.3/93.7$ & $183/121/144$ \\
        \bottomrule
    \end{tabular}
    \vspace{-0.25cm}
\end{table*}

\begin{table*}[p]
    \centering
    \footnotesize
    \setlength{\tabcolsep}{5pt}
    \caption{Per-task summary of the golden-ticket vs.\ Gaussian comparison across all $51$ task--policy pairs. \textbf{Base} and \textbf{Golden} are held-out success rates (\%); \textbf{Search} combines the per-task budget (tickets $\times$ search-envs per ticket) with the search method (RS = random search, CEM = cross-entropy method). Shaded \emph{Avg.} rows give the per-benchmark mean across the listed tasks.}
    \label{tab:existence}
    \begin{tabular}{@{}llrrl@{}}
        \toprule
        \textbf{Benchmark} & \textbf{Task} & \textbf{Base} & \textbf{Golden} & \textbf{Search} \\
        \midrule
        \rowcolor{gray!12} franka\_sim & Avg.\ over $4$ ckpts & $38.5 \pm 27.9$ & $96.0 \pm 2.4$ & ${\approx}250 \times 25$ (RS) \\
        \midrule
        \multirow[t]{5}{*}{robomimic} & \texttt{Lift} & $78.4 \pm 3.9$ & $96.2 \pm 1.4$ & \multirow[t]{5}{*}{$5000 \times 100$ (RS/CEM)} \\
         & \texttt{Can} & $42.8 \pm 4.4$ & $95.0 \pm 2.1$ & \\
         & \texttt{Square} & $52.2 \pm 3.6$ & $42.6 \pm 5.7$ & \\
         & \texttt{Transport} & $10.7 \pm 2.2$ & $18.8 \pm 3.6$ & \\
        \rowcolor{gray!12} & Avg. & $46.0$ & $63.1$ & \\
        \midrule
        \multirow[t]{11}{*}{LIBERO-Object} & T0 & $82$ & $100$ & \multirow[t]{11}{*}{$1416 \times 5$ (RS)} \\
         & T1 & $98$ & $100$ & \\
         & T2 & $98$ & $100$ & \\
         & T3 & $98$ & $100$ & \\
         & T4 & $76$ & $100$ & \\
         & T5 & $82$ & $100$ & \\
         & T6 & $100$ & $100$ & \\
         & T7 & $90$ & $100$ & \\
         & T8 & $96$ & $100$ & \\
         & T9 & $100$ & $100$ & \\
        \rowcolor{gray!12} & Avg. & $92.0$ & $100.0$ & \\
        \midrule
        \multirow[t]{11}{*}{LIBERO-Goal} & T0 & $72$ & $100$ & \multirow[t]{11}{*}{$1338 \times 3$--$5$ (RS)} \\
         & T1 & $94$ & $100$ & \\
         & T2 & $86$ & $100$ & \\
         & T3 & $52$ & $68$ & \\
         & T4 & $92$ & $100$ & \\
         & T5 & $78$ & $100$ & \\
         & T6 & $80$ & $94$ & \\
         & T7 & $100$ & $100$ & \\
         & T8 & $94$ & $100$ & \\
         & T9 & $68$ & $90$ & \\
        \rowcolor{gray!12} & Avg. & $81.6$ & $95.2$ & \\
        \midrule
        \multirow[t]{11}{*}{LIBERO-Spatial} & T0 & $78$ & $86$ & \multirow[t]{11}{*}{$1081 \times 3$--$5$ (RS)} \\
         & T1 & $94$ & $98$ & \\
         & T2 & $84$ & $94$ & \\
         & T3 & $62$ & $100$ & \\
         & T4 & $84$ & $92$ & \\
         & T5 & $58$ & $90$ & \\
         & T6 & $90$ & $100$ & \\
         & T7 & $90$ & $96$ & \\
         & T8 & $78$ & $90$ & \\
         & T9 & $86$ & $92$ & \\
        \rowcolor{gray!12} & Avg. & $80.4$ & $93.8$ & \\
        \midrule
        \multirow[t]{8}{*}{SimplerEnv (WidowX)} & \texttt{close\_drawer} & $65$ & $100.0$ & \multirow[t]{8}{*}{$250 \times (5\text{--}10)$ (CEM)} \\
         & \texttt{open\_drawer} & $84$ & $97.3$ & \\
         & \texttt{put\_eggplant\_in\_basket} & $63$ & $96.0$ & \\
         & \texttt{put\_eggplant\_in\_sink} & $21$ & $100.0$ & \\
         & \texttt{stack\_cube} & $54$ & $77.0$ & \\
         & \texttt{carrot\_on\_plate} & $72$ & $93.0$ & \\
         & \texttt{spoon\_on\_towel} & $82$ & $90.7$ & \\
        \rowcolor{gray!12} & Avg. & $63.0$ & $93.4$ & \\
        \midrule
        \multirow[t]{6}{*}{DexMimicGen} & \texttt{Drawer Cleanup} & $82.2 \pm 4.4$ & $88.0 \pm 2.9$ & \multirow[t]{6}{*}{$500 \times 50$ (RS)} \\
         & \texttt{Tray Lift} & $71.0 \pm 3.9$ & $79.9 \pm 4.6$ & \\
         & \texttt{Threading} & $62.0 \pm 4.1$ & $60.0 \pm 4.9$ & \\
         & \texttt{Box Cleanup} & $87.6 \pm 2.4$ & $97.8 \pm 1.1$ & \\
         & \texttt{Piece Assembly} & $68.3 \pm 3.3$ & $75.6 \pm 3.1$ & \\
        \rowcolor{gray!12} & Avg. & $74.2$ & $80.3$ & \\
        \midrule
        \multirow[t]{5}{*}{Franka HW} & \texttt{cube\_pick} & $80$ & $98$ & $6 \times 25$ (RS) \\
         & \texttt{piston\_assembly} & $60$ & $88$ & $22$ tickets, $61$ eps (RS) \\
         & \texttt{banana\_pick} & $50$ & $68$ & $10 \times 5$ (RS) \\
         & \texttt{cup\_push} & $40$ & $100$ & $10 \times 5$ (RS) \\
        \rowcolor{gray!12} & Avg. & $57.5$ & $88.5$ & --- \\
        \bottomrule
    \end{tabular}
\end{table*}

\section{Statistical Significance of the Improvement Rate}
\label{app:significance}
Throughout, \emph{base} denotes the held-out success rate of the pretrained policy with Gaussian initial noise, and \emph{gold} the held-out success rate of the best searched constant initial noise (golden ticket) for the same task--policy pair. The suite comprises $n=51$ pairs: franka\_sim ($1$, mean of $4$ seeds), robomimic ($4$), LIBERO ($30$), DexMimicGen ($5$), SimplerEnv (WidowX) ($7$), and real Franka hardware ($4$).

For each of the $n=51$ tasks $i$, we define a binary outcome $X_i=1$ if the best searched noise vector strictly outperforms the base (Gaussian noise) policy on the held-out evaluation, and $X_i=0$ otherwise. Modeling $X_i \stackrel{\text{iid}}{\sim} \mathrm{Bernoulli}(p)$, we test
\[
    H_0\!:\ p \le 0.5 \quad\text{vs.}\quad H_1\!:\ p > 0.5,
\]
where $p$ is the probability that, for a task drawn from the distribution sampled by our benchmark suite (franka\_sim $\cup$ LIBERO $\cup$ robomimic $\cup$ DexMimicGen $\cup$ SimplerEnv (WidowX) $\cup$ real-world), the best searched noise vector strictly improves over Gaussian sampling. The null $p_0=\tfrac{1}{2}$ corresponds to the standard sign-test null.

Letting $K=\sum_i X_i$, the exact one-sided p-value is $\mathbb{P}_{p=0.5}(K\ge k_{\mathrm{obs}})=2^{-n}\sum_{j=k_{\mathrm{obs}}}^{n}\binom{n}{j}$. For $k=46$ we obtain p-value $\approx 1.16\times 10^{-9}$, and for the weaker ``at-least-match'' count $k=49$, p-value $\approx 5.89\times 10^{-13}$. We report a Wilson 95\% score interval for $p$,
\[
    \mathrm{CI} = \frac{\hat p + \tfrac{z^2}{2n} \pm z\sqrt{\tfrac{\hat p(1-\hat p)}{n} + \tfrac{z^2}{4n^2}}}{1 + z^2/n}, \qquad z=1.96,
\]
which gives $[0.79,0.96]$ for $\hat p=46/51$ and $[0.87,0.99]$ for $\hat p=49/51$.

\textit{Approximations.} (i) Tasks within a benchmark may share a base policy and also the observation pipeline (e.g., all $30$ LIBERO tasks use SmolVLA), so the iid assumption is approximate. (ii) The inference applies to the population sampled by our benchmarks, not to all manipulation tasks. (iii) The indicator $X_i$ does not weight by the per-task uncertainty in the success-rate estimate.

\section{Performance Improvement with Golden Tickets}
\label{app:reach}
Beyond whether golden tickets improve over Gaussian sampling (Appendix~\ref{app:significance}), we ask: \emph{above what base policy performance do golden tickets reliably improve task success?} Concretely, we seek the lowest base policy floor $T^\star$ such that, among the task--policy pairs whose base success rate is at least $T^\star$, at least $90\%$ are pushed past a high absolute success bar by their golden ticket. We instantiate ``high'' at two levels, $G \in \{90, 97\}\%$.

For a base floor $T$ and absolute bar $G$, let
\[
    \hat p(T, G) \;=\; \frac{\#\{\text{pairs with } \mathrm{base} \ge T \text{ and } \mathrm{gold} \ge G\}}{\#\{\text{pairs with } \mathrm{base} \ge T\}}
\]
be the empirical pass rate among the pairs included by floor $T$. We then report
\[
    T^\star_G \;=\; \min\bigl\{\, T : \hat p(T^\prime, G) \ge 0.90 \ \text{ for every } T^\prime \ge T \,\bigr\},
\]
i.e., the lowest base floor above which the empirical pass rate \emph{stays} at $\ge 90\%$. Candidate floors are the distinct base success rates in the suite. Table~\ref{tab:reach} reports $T^\star_G$ for both bars.

\begin{center}
    \small
    \captionof{table}{Base policy floor $T^\star_G$ above which golden tickets reach the absolute bar $G$ in at least $90\%$ of the supported task--policy pairs. $n$ is the number of pairs with base $\ge T^\star_G$; $k$ is the number of those reaching $\ge G$.}
    \label{tab:reach}
    \begin{tabular}{ccccc}
        \toprule
        Target $G$ & $T^\star_G$ & Support $n$ & Successes $k$ & $k/n$   \\
        \midrule
        $90\%$     & $71\%$      & $34$        & $31$          & $0.912$ \\
        $97\%$     & $87.6\%$    & $15$        & $14$          & $0.933$ \\
        \bottomrule
    \end{tabular}
\end{center}

\paragraph{Base floor $T^\star=71\%$ (bar $G=90\%$, $n=34$).} The $34$ pairs with base $\ge 71\%$ are (including $3$ trivial passes where the base policy is already at $100\%$):
\begin{itemize}
    \setlength\itemsep{0pt}
    \item LIBERO-Spatial ($8$): \texttt{T0}, \texttt{T1}, \texttt{T2}, \texttt{T4}, \texttt{T6}, \texttt{T7}, \texttt{T8}, \texttt{T9}\quad(excludes \texttt{T3}, \texttt{T5})
    \item LIBERO-Goal ($8$): \texttt{T0}, \texttt{T1}, \texttt{T2}, \texttt{T4}, \texttt{T5}, \texttt{T6}, \texttt{T7}, \texttt{T8}\quad(excludes \texttt{T3}, \texttt{T9})
    \item LIBERO-Object ($10$): all tasks \texttt{T0}--\texttt{T9}
    \item DexMimicGen ($3$): \texttt{Drawer Cleanup}, \texttt{Tray Lift}, \texttt{Box Cleanup}
    \item robomimic ($1$): \texttt{Lift}
    \item Real robot ($1$): cube picking
    \item SimplerEnv (WidowX) ($3$): \texttt{carrot\_on\_plate}, \texttt{spoon\_on\_towel}, \texttt{open\_drawer}
\end{itemize}
Three of these $34$ pairs do not reach $\ge 90\%$:
\begin{itemize}
    \setlength\itemsep{0pt}
    \item DexMimicGen \texttt{Tray Lift}: base $71\%$, gold $79.9\%$
    \item DexMimicGen \texttt{Drawer Cleanup}: base $82.2\%$, gold $88\%$
    \item LIBERO-Spatial \texttt{T0}: base $78\%$, gold $86\%$
\end{itemize}

\paragraph{Base floor $T^\star=87.6\%$ (bar $G=97\%$, $n=15$).} The $15$ pairs with base $\ge 87.6\%$ are:
\begin{itemize}
    \setlength\itemsep{0pt}
    \item LIBERO-Spatial ($3$): \texttt{T1}, \texttt{T6}, \texttt{T7}
    \item LIBERO-Goal ($4$): \texttt{T1}, \texttt{T4}, \texttt{T7}, \texttt{T8}
    \item LIBERO-Object ($7$): \texttt{T1}, \texttt{T2}, \texttt{T3}, \texttt{T6}, \texttt{T7}, \texttt{T8}, \texttt{T9}
    \item DexMimicGen ($1$): \texttt{Box Cleanup}
\end{itemize}
One of these $15$ pairs does not reach $\ge 97\%$:
\begin{itemize}
    \setlength\itemsep{0pt}
    \item LIBERO-Spatial \texttt{T7}: base $90\%$, gold $96\%$
\end{itemize}

\section{Further Results}
\label{app:dsrl}
\subsection{DexMimicGen Results}
\label{app:dexmimicgen-results}
\begin{figure}[t]
    \centering
    \begin{subfigure}[b]{\linewidth}
        \centering
        \includegraphics[width=\linewidth]{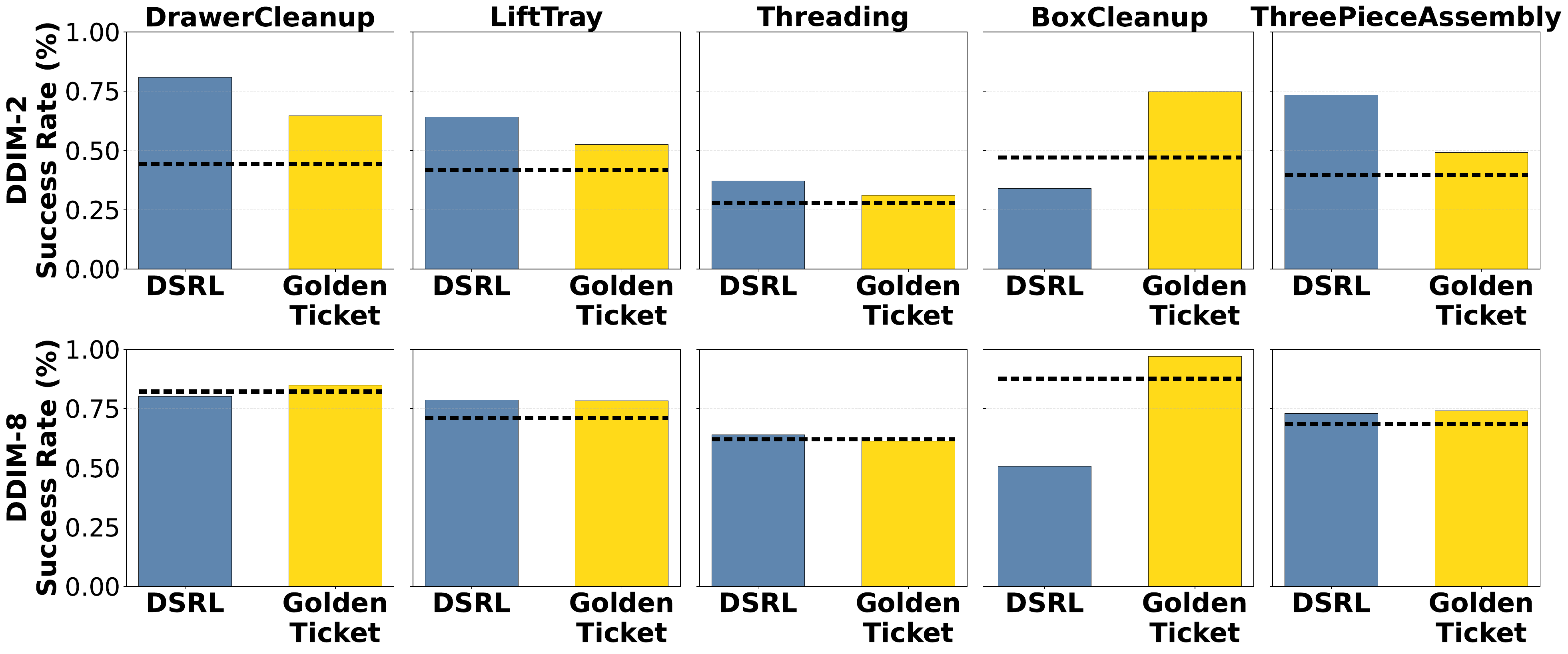}
        \caption{$5$k search episodes per task.}
        \label{fig:dsrl-budget-5k}
    \end{subfigure}
    \vspace{4pt}
    \begin{subfigure}[b]{\linewidth}
        \centering
        \includegraphics[width=\linewidth]{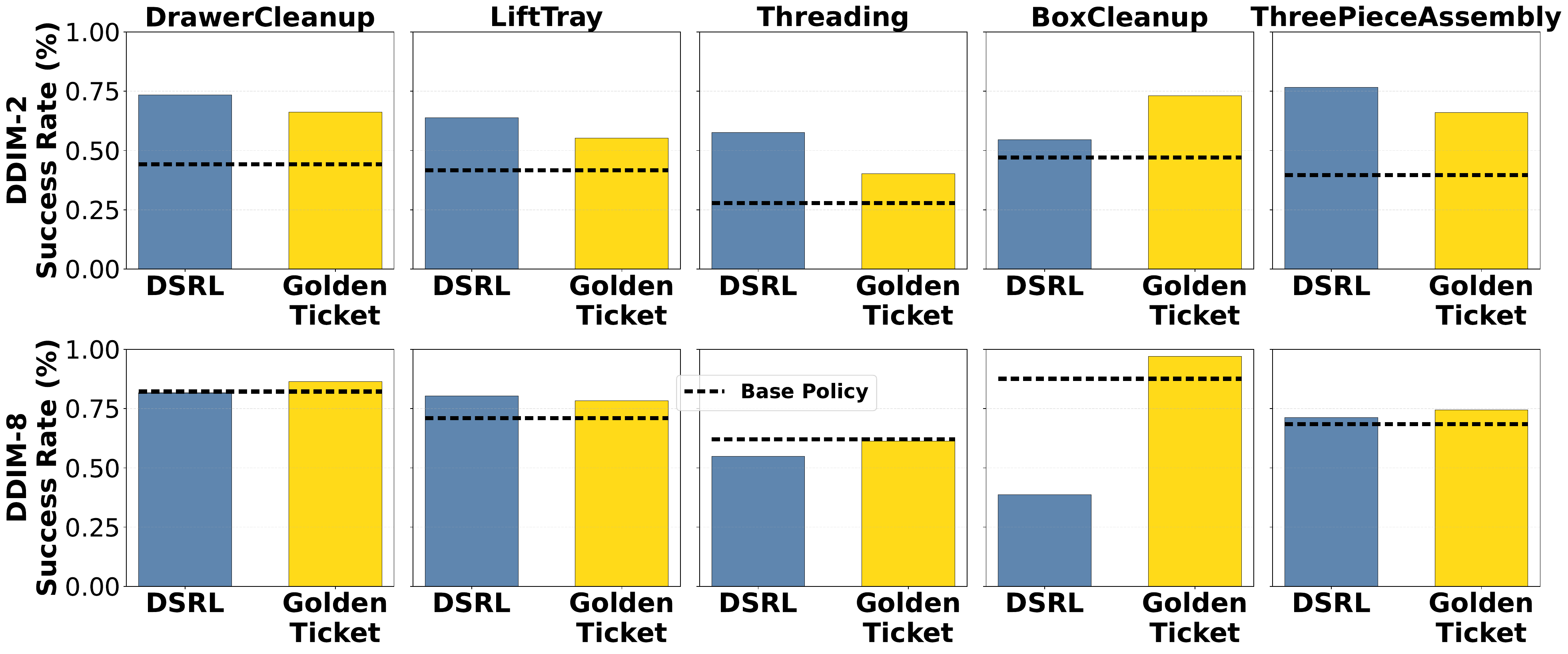}
        \caption{$10$k search episodes per task.}
        \label{fig:dsrl-budget-10k}
    \end{subfigure}
    \caption{DSRL (\textbf{\textcolor{profblue}{blue}}) vs.\ Golden Ticket via Random Search (\textbf{\textcolor{profgold}{yellow}}) on DexMimicGen across five tasks (columns), two DDIM step settings (top row of each panel: DDIM-$2$; bottom row: DDIM-$8$), and two search-episode budgets ($5$k in (a), $10$k in (b)). The dashed horizontal line in each panel marks the base-policy success rate at that DDIM setting.}
    \label{fig:dsrl-budget-grid}
    \vspace{-0.5cm}
\end{figure}

Figure~\ref{fig:dsrl-budget-grid} reports per-task success rate for DSRL and GT Search(Random Search) at two episode budgets ($5$k and $10$k search episodes per task) crossed with two DDIM step counts ($2$ and $8$); the dashed line in each panel marks the corresponding base-policy success rate. The five tasks are from the DexMimicGen suite: \texttt{DrawerCleanup}, \texttt{LiftTray}, \texttt{Threading}, \texttt{BoxCleanup}, and \texttt{ThreePieceAssembly}. GT Search appears broadly equivalent to DSRL in sample efficiency on DexMimicGen. DSRL often struggles to match base-policy performance at DDIM-$8$, while doing better at DDIM-$2$. The standout task is \texttt{BoxCleanup}, where GT Search clearly outperform DSRL across DDIM steps.

\subsection{Multi-task Results}
\label{app:multi-task}
\subsubsection{Is there evidence of golden tickets that generalize across tasks?}
In LIBERO with SmolVLA across the $3$ task suites (LIBERO-Object, LIBERO-Goal, LIBERO-Spatial; $10$ tasks each, $30$ tasks total), we investigate whether golden tickets found (using random search) for one task can also act as golden tickets for other tasks.
Across all $30$ tasks, golden tickets match or exceed the base policy's performance (Table~\ref{tab:libero_tickets}). Certain golden tickets match or exceed the base policy on multiple tasks: in LIBERO-Goal, ticket \texttt{\#03c2} matches or exceeds the base policy in $7/10$ tasks (\texttt{T0}, \texttt{T1}, \texttt{T2}, \texttt{T4}, \texttt{T6}, \texttt{T7}, \texttt{T8}); in LIBERO-Object, ticket \texttt{\#015a} matches or exceeds the base policy in $7/10$ tasks (\texttt{T2}, \texttt{T3}, \texttt{T5}, \texttt{T6}, \texttt{T7}, \texttt{T8}, \texttt{T9}); in LIBERO-Spatial, ticket \texttt{\#60c0} matches or exceeds the base policy in $5/10$ tasks (\texttt{T0}, \texttt{T2}, \texttt{T3}, \texttt{T6}, \texttt{T8}).
Our results provide evidence that related tasks may share golden tickets for VLA policies.

\subsubsection{Multi-task Golden Tickets Search}
We complement the per-task SimplerEnv results in Appendix~\ref{app:existence} with multi-task golden ticket searches on the GR00T~N1.5 base policy, run separately on a drawer pair (\texttt{open\_drawer}, \texttt{close\_drawer}) and an eggplant pair (\texttt{put\_eggplant\_in\_sink}, \texttt{put\_eggplant\_in\_basket}). Each search uses CEM with population $20$ for up to $25$ iterations and a maximum of $500$ tickets, with $10$ search-environment instances per task ($20$ envs total) and the task-averaged search-environment return as the objective; the top-ranked tickets returned by CEM are then re-evaluated on $300$ held-out episodes per task (Table~\ref{tab:simpler_widowx_multitask_2task}). All four tickets improve on the base policy on both tasks of their pair: the drawer tickets lift the $2$-task average from $74.5$ to up to $97.8$, and the eggplant tickets from $42.0$ to up to $74.5$.

\begin{table}[t]
    \centering
    \footnotesize
    \caption{Success rates (\%) for base policy and lottery tickets on LIBERO-Goal, LIBERO-Object, and LIBERO-Spatial with SmolVLA. T0 - T9 represent different tasks. Top-1 ticket per task shown (unique set) plus best average ticket ($\star$). Underline: $\geq$ base policy performance.}
    \label{tab:libero_tickets}
    \setlength{\tabcolsep}{2pt}
        \resizebox{0.86\linewidth}{!}{
            \begin{tabular}{l cccccccccc c}
                \toprule
                \multicolumn{12}{c}{\textbf{LIBERO-Goal}}                                                                                                                                                                                               \\
                \textbf{\#Ticket}
                               & \textbf{T0}    & \textbf{T1}              & \textbf{T2}    & \textbf{T3} & \textbf{T4}
                               & \textbf{T5}    & \textbf{T6}              & \textbf{T7}    & \textbf{T8} & \textbf{T9}
                               & \textbf{Avg}                                                                                                                                                                                                           \\
                \midrule

                \rowcolor{gray!20}
                Base           & 72             & 94                       & 86             & 52          & 92                       & 78           & 80                      & \textbf{100}             & 94             & 68          & \textbf{81.6} \\

                \#03c2 $\star$ & \underline{84} & \underline{\textbf{100}} & \underline{92} & 40          & \underline{\textbf{100}} & 64           & \underline{\textbf{94}} & \underline{\textbf{100}} & \underline{98} & 18          & 79.0          \\
                \#2cee         & 78             & 84                       & \textbf{100}   & 2           & 98                       & 76           & 22                      & \textbf{100}             & 86             & 0           & 64.6          \\
                \#1ff8         & 98             & 64                       & 86             & 14          & \textbf{100}             & 0            & 38                      & 96                       & \textbf{100}   & 2           & 59.8          \\
                \#4d97         & 2              & 96                       & 16             & \textbf{68} & 98                       & 92           & 30                      & 54                       & \textbf{100}   & 4           & 56.0          \\
                \#672f         & 0              & 12                       & 88             & 0           & 72                       & 4            & 38                      & \textbf{100}             & 2              & \textbf{90} & 40.6          \\
                \#0f3f         & \textbf{100}   & 2                        & 0              & 0           & 68                       & 56           & 0                       & 76                       & 38             & 0           & 34.0          \\
                \#0310         & 0              & 12                       & 0              & 0           & \textbf{100}             & \textbf{100} & 10                      & 4                        & 42             & 0           & 26.8          \\

                \bottomrule
            \end{tabular}
        }

        \vspace{0.4em}

        \resizebox{0.86\linewidth}{!}{
            \begin{tabular}{l cccccccccc c}
                \toprule
                \multicolumn{12}{c}{\textbf{LIBERO-Object}}                                                                                                                                                                                                          \\
                \textbf{\#Ticket}
                               & \textbf{T0}  & \textbf{T1}  & \textbf{T2}              & \textbf{T3}              & \textbf{T4}
                               & \textbf{T5}  & \textbf{T6}  & \textbf{T7}              & \textbf{T8}              & \textbf{T9}
                               & \textbf{Avg}                                                                                                                                                                                                                        \\
                \midrule

                \rowcolor{gray!20}
                Base           & 82           & 98           & 98                       & 98                       & 76           & 82             & \textbf{100}             & 90             & 96                       & \textbf{100}             & \textbf{92.0} \\

                \#015a $\star$ & 36           & 62           & \underline{\textbf{100}} & \underline{\textbf{100}} & 16           & \underline{92} & \underline{\textbf{100}} & \underline{94} & \underline{\textbf{100}} & \underline{\textbf{100}} & 80.0          \\
                \#5159         & 20           & 86           & \textbf{100}             & \textbf{100}             & 60           & \textbf{100}   & 94                       & 36             & 92                       & \textbf{100}             & 78.8          \\
                \#184f         & 98           & 0            & 94                       & 86                       & \textbf{100} & 0              & 78                       & \textbf{100}   & 86                       & 44                       & 68.6          \\
                \#1944         & \textbf{100} & 38           & 72                       & 94                       & 94           & 4              & 30                       & 90             & 72                       & 0                        & 59.4          \\
                \#14e7         & 14           & \textbf{100} & 0                        & 0                        & 0            & 32             & 56                       & 64             & 56                       & 0                        & 32.2          \\
                \#096d         & 50           & 0            & 0                        & 4                        & 0            & 0              & 0                        & \textbf{100}   & 0                        & 0                        & 15.4          \\

                \bottomrule
            \end{tabular}
        }

        \vspace{0.4em}

        \resizebox{0.86\linewidth}{!}{
            \begin{tabular}{l cccccccccc c}
                \toprule
                \multicolumn{12}{c}{\textbf{LIBERO-Spatial}}                                                                                                                                                        \\
                \textbf{\#Ticket}
                               & \textbf{T0}    & \textbf{T1} & \textbf{T2}    & \textbf{T3}              & \textbf{T4}
                               & \textbf{T5}    & \textbf{T6} & \textbf{T7}    & \textbf{T8}              & \textbf{T9}
                               & \textbf{Avg}                                                                                                                                                                       \\
                \midrule

                \rowcolor{gray!20}
                Base           & 78             & 94          & 84             & 62                       & 84          & 58          & 90             & 90          & 78             & 86          & \textbf{80.4} \\

                \#60c0 $\star$ & \underline{82} & 82          & \underline{88} & \underline{\textbf{100}} & 74          & 24          & \underline{98} & 80          & \underline{82} & 68          & 77.8          \\
                \#a68f         & 76             & 72          & 92             & 50                       & \textbf{92} & \textbf{90} & 56             & 70          & \textbf{90}    & 82          & 77.0          \\
                \#ecf0         & 68             & 96          & 80             & 76                       & 12          & 30          & 98             & \textbf{96} & 82             & 68          & 70.6          \\
                \#0e0c         & 80             & 60          & 86             & \textbf{100}             & 10          & 24          & 90             & 92          & 84             & 74          & 70.0          \\
                \#05c7         & 70             & 92          & 88             & 72                       & 52          & 0           & \textbf{100}   & 64          & 44             & 78          & 66.0          \\
                \#517b         & 72             & 84          & \textbf{92}    & 78                       & 2           & 20          & 66             & 92          & 62             & 52          & 62.0          \\
                \#6a21         & 64             & \textbf{98} & 90             & 30                       & \textbf{92} & 30          & 30             & 74          & 58             & 44          & 61.0          \\
                \#7c0e         & 64             & 68          & 64             & 4                        & 22          & 60          & 6              & 88          & 44             & \textbf{92} & 51.2          \\
                \#3b9d         & \textbf{84}    & 36          & 46             & 94                       & 14          & 6           & 88             & 22          & 16             & 0           & 40.6          \\

                \bottomrule
            \end{tabular}
        }
\end{table}

\begin{table}[t]
    \centering
    \footnotesize
    \setlength{\tabcolsep}{6pt}
    \caption{Multi-task golden ticket search on two SimplerEnv (WidowX) task pairs. Cells are held-out success rates (\%, $300$ episodes per task); \textbf{Avg.} is the pair average.}
    \label{tab:simpler_widowx_multitask_2task}
    \setlength{\tabcolsep}{4pt}
    \begin{tabular}{@{}l l c c c c@{}}
        \toprule
        \textbf{Task pair} & \textbf{Ticket} & \textbf{Search idx} & \textbf{Task $1$} & \textbf{Task $2$} & \textbf{Avg.} \\
        \midrule
        \multirow{3}{*}{\shortstack[l]{Drawer\\\texttt{OD}, \texttt{CD}}}     & Base policy     & ---                 & $84$              & $65$              & $74.5$ \\
                                                              & ticket\#$1$     & $201$               & $90.0$            & $\mathbf{98.3}$   & $94.2$ \\
                                                              & ticket\#$2$     & $225$               & $\mathbf{97.3}$   & $\mathbf{98.3}$   & $\mathbf{97.8}$ \\
        \midrule
        \multirow{3}{*}{\shortstack[l]{Eggplant\\\texttt{PES}, \texttt{PEB}}} & Base policy     & ---                 & $21$              & $63$              & $42.0$ \\
                                                              & ticket\#$1$     & $382$               & $60.7$            & $\mathbf{88.3}$   & $\mathbf{74.5}$ \\
                                                              & ticket\#$2$     & $410$               & $\mathbf{64.0}$   & $82.3$            & $73.2$ \\
        \bottomrule
    \end{tabular}
    \vspace{-0.25cm}
\end{table}

\end{document}